%% file: main.tex
\pdfoutput=1
\documentclass{article}

\PassOptionsToPackage{numbers, sort}{natbib}
\makeatletter
\def\input@path{{templates/}}
\makeatother
\usepackage[main, final]{neurips_2026}
\usepackage[utf8]{inputenc}
\usepackage[T1]{fontenc}
\usepackage{hyperref}
\usepackage{url}
\usepackage{booktabs}
\usepackage{nicefrac}
\usepackage{microtype}
\usepackage{xcolor}
\definecolor{fancyColor}{HTML}{7E5A83}
\usepackage{mathtools}
\usepackage{tikz}
\usetikzlibrary{arrows.meta,calc,positioning,fit}
\usepackage{nanostyle}

\title{Tropical Gaussian Anticoncentration:\\Optimal Instance-Dependent Bounds for\\Online Learning in Extensive-Form Games}

\author{%
  Ashkan Soleymani$^*$ \\
  MIT \\
  \texttt{ashkanso@mit.edu} \\
  \And
  Zhiyuan Fan$^*$ \\
  MIT \\
  \texttt{fanzy@mit.edu} \\
  \And
  Lillian J.~Ratliff \\
  University of Washington \\
  \texttt{ratliffl@uw.edu} \\
  \And
  Patrick Jaillet \\
  MIT \\
  \texttt{jaillet@mit.edu} \\
  \And
  Gabriele Farina \\
  MIT \\
  \texttt{gfarina@mit.edu}
}

\newcommand{\xb}{\bm{x}}
\newcommand{\yb}{\bm{y}}
\newcommand{\ub}{\bm{u}}
\newcommand{\wb}{\bm{w}}
\newcommand{\rb}{\bm{r}}
\newcommand{\gb}{\bm{g}}

\newcommand{\seqs}{\Sigma}
\newcommand{\nonrootseqs}{\Sigma^+}
\newcommand{\leaveseqs}{\cE}
\newcommand{\nonleaveseqs}{\Sigma \setminus \cE}
\newcommand{\tfsdp}{\cH}
\newcommand{\DilEnt}{\varphi_{\mathrm{DE}}}
\newcommand{\Qwidth}{\|\cQ\|_{\perp}}

\newcommand{\Ynorm}{_{\tfsdp,1}}
\newcommand{\Qnorm}{_{\tfsdp,\infty}}

\newcommand{\Heff}{H_{\mathrm{eff}}}

\begin{document}

\maketitle

{\let\thefootnote\relax\footnotetext{$^*$ Equal contribution}}

\begin{abstract}
We study the minimax regret of full-information online decision making in extensive-form games. For prediction with expert advice, the optimal regret is $\Theta(\sqrt{T\log K})$, where $K$ is the number of experts, or equivalently, the number of normal-form actions.
For treeplex strategy spaces, existing efficient algorithms, including KOMWU and online mirror descent with the weight-one dilated entropy regularizer, achieve regret $\mathcal{O}(\sqrt{T\log|\mathcal{V}|})$, where $\mathcal{V}$ is the set of reduced normal-form strategies. We prove that this logarithmic dependence is minimax optimal in the worst-case full-information reward model.
The main difficulty is that pure plans in a treeplex are not independent experts: their rewards are correlated through the recursive structure of decisions and observations. To capture this correlation, we introduce tropical Gaussian formulae, recursive Gaussian expressions built from maxima and normalized sums which preserve the variance.
The index count of such a formula matches the number of reduced normal-form continuation plans in the corresponding subtree. Our main analytic result shows that every log-balanced tropical Gaussian formula with index count $N$ has Gaussian width $\Omega(\sqrt{\log N})$. 
While a direct approach struggles to recover this lower bound, our proof uses a spherical entropy profile to measure scale-dependent separation among the Gaussian directions, and then applies Tur\'an's theorem and Sudakov minoration to obtain the width lower bound.
Applying this result to treeplexes gives an oblivious hard distribution with valid transition outcome and Rademacher terminal rewards, establishing a regret lower bound of $\Omega(\sqrt{T\log|\mathcal{V}|})$. This matches known upper bounds up to universal constants and shows that the DilEnt/KOMWU dependence on the reduced normal-form complexity is unavoidable.
\end{abstract}

\section{Introduction}

Determining the minimax regret bound is a central problem in online learning. One of the most fundamental instances is prediction with expert advice, where a player selects a probability distribution over a set of experts $\{1,2,\ldots,K\}$ in each round. After making the choice, the learner observes the rewards of all experts, which may be assigned adversarially within $[-1,1]$. The goal is to minimize the \emph{cumulative regret}, defined as the reward of the best expert in hindsight minus the learner's expected total reward after $T$ rounds.

The lower bound is almost immediate in this normal-form problem: each expert is assigned an independent Rademacher reward of $-1$ or $1$ in every round. The player suffers a regret of $\Theta(\sqrt{T\log K})$, since it gains no advantage from historical data, while the best fixed expert is given by the maximum of $K$ random walks, with expected value $\Theta(\sqrt{T\log K})$ \citep{cesa2006prediction}. This regret bound is tight, as it matches the regret upper bound given by Hedge~\citep{freund1997decision}, also known as multiplicative weights update.

A natural generalization of this normal-form problem is that of extensive-form games (EFGs), in which, at each round, the player needs to act sequentially in a multi-stage setting. The player's behavioral strategy space can be characterized by a sequential decision problem \citep{von1996efficient}; geometrically, it is a recursively assembled treeplex. The reduced normal-form set $\cV$ encodes the set of pure plans, which can be exponentially large compared to the sequence-form description. \citet{farina2022kernelized} show that KOMWU achieves $\cO(\sqrt{T\log|\cV|})$ regret while remaining computationally efficient. It is shown to be iterate-equivalent to online mirror descent with weight-one dilated entropy (DilEnt) \citep{bai2022efficient}, and \citet{fan2024optimality} recover the same bound through treeplex norms. However, none of these works shows that this regret bound is minimax optimal.

As a result, it is not clear a priori whether the \(\log|\cV|\) scale is an intrinsic minimax dependence, analogous to the \(\log K\) term in prediction with experts, or merely an artifact of existing analyses. The obstacle to reusing the classical experts lower bound is correlation. A reduced-normal-form pure plan chooses one action at each decision point in its continuation support, and after an observation it specifies continuations at all successor information sets. Thus pure-plan rewards cannot be modeled as independent experts, but their correlations are not arbitrary either: they are constrained by the recursive decision-and-observation structure of the treeplex.

Our hard instance keeps the randomness as elementary as possible. The adversary is oblivious, reward randomness is independent across disjoint action subtrees, terminal rewards are Rademacher, and observation outcomes are routed independently across rounds using log-balanced probabilities. Under this construction, the covariance-matched comparator process has a tropical-Gaussian recursion: decision nodes produce maxima over action subtrees, while observation nodes produce variance-normalized sums of the successor processes. We visualize the connection between these two concepts in \Cref{fig:efg-formula-bridge}. The main analytic difficulty is to prove that these correlated max-linear Gaussian expressions still have Gaussian width of order \(\sqrt{\log|\cV|}\). A finite-time high-dimensional Gaussian comparison then transfers this Gaussian width lower bound back to the \(T\)-round Rademacher hard instance, showing that every online learner suffers expected regret
\[
    \Omega(\sqrt{T\log|\cV|}).
\]
in EFGs when \(T\ge\Omega(\log^{14}|\cV|)\). This shows that the \(\log |\cV|\) dependence appearing in KOMWU and DilEnt characterizes the intrinsic worst-case minimax quantity of the problem.

\begin{figure}[t]
    \centering
    \resizebox{\linewidth}{!}{\input{figures/efg_formula_bridge}}
    \refstepcounter{figure}
    \label{fig:efg-formula-bridge}
    \vspace{6pt}
    \parbox{\linewidth}{\small\textbf{Figure~\thefigure:} A concrete treeplex example and its tropical Gaussian formula. The root decision gives the disjoint union \(\{L\}\cup(\cV_{j_1}\times\cV_{j_2})\), and the observation after \(R\) makes each pure plan specify one continuation in both successor subtrees. Thus the five pure-plan vertices index the five Gaussian directions in the expanded maximum; for example, \((R,A,D)\) indexes \((G_A+G_D)/\sqrt2\).}
    \vspace{-12pt}
\end{figure}

The main technical challenge is correlation: the terms in the normal form of the Gaussian process can be highly correlated, so the count \(N_Z\) alone does not imply the desired width. Instead, we introduce a spherical entropy profile to characterize this correlation. This metric measures how many pairs of directions are effectively close and can be computed inductively in a bottom-up manner. An upper bound on this correlation quantity allows us to apply Turán's theorem to extract a family of well-separated vectors, which fits into Sudakov minoration to lower bound the Gaussian width.

Our contributions are as follows.
\begin{itemize}[itemsep=0pt, topsep=0pt, leftmargin=*]
    \item We reduce full-information online decision making on a treeplex to tropical Gaussian formulae. The hard instance is an oblivious construction with transition outcomes and Rademacher terminal rewards, and it preserves the treeplex recursion exactly.

    \item We introduce a spherical-entropy method for the correlated pure-plan reward processes generated by the reduction. The resulting formula lower bound gives a regret lower bound of \(\Omega(\sqrt{T\log|\cV|})\) for the worst-case EFG full-information reward model, matching the upper bounds achieved by KOMWU and online mirror descent with the DilEnt regularizer.

    \item We show that weight-one DilEnt and KOMWU are minimax optimal on every treeplex. Under norms whose dual controls all feasible reward vectors, no such OMD guarantee can improve the \(\log|\cV|\) dependence, while weight-one DilEnt attains it up to universal constants.
\end{itemize}

\section{Related Work}

\paragraph{Equilibrium computation in general-sum EFGs.}
Polynomial-time equilibrium computation in compact game representations was first approached through linear programming formulations \citep{papadimitriou2008computing,jiang2011polynomial}. These formulations are important structurally, but their solver dependence is not the preferred route for large sequential games. Modern scalable methods are instead built around regret minimization and first-order dynamics. \citet{syrgkanis2015fast} introduced the RVU condition for optimistic no-regret algorithms and used it to obtain \(T^{1/4}\) individual regret, hence a \(T^{-3/4}\) rate to coarse correlated equilibrium (CCE). \citet{daskalakis2021near} later proved polylogarithmic regret in general games, implying a \(\widetilde{\cO}(1/T)\) CCE rate, and \citet{farina2022kernelized} extended this fast-rate picture to \(0/1\)-polyhedral games, including EFGs. A parallel line studies clairvoyant dynamics: \citet{piliouras2022beyond} introduced Clairvoyant MWU, and \citet{farina2022clairvoyant} related clairvoyant regret minimization to conceptual prox methods from first-order optimization \citep{chen1993convergence,nemirovski2004prox}. Other near-\(1/T\) rates for general convex games were obtained by \citet{farina2022near}. More recently, \citet{soleymani2025faster,soleymani2025cautious} introduced cautious optimism, a meta-algorithm that accelerates no-regret learning dynamics in general convex games to a \(\widetilde{\cO}(1/T)\) rate.
Counterfactual regret minimization remains a central EFG-specific framework \citep{zinkevich2007regret}; its classical regret scales as \(\sqrt{T}\), while optimistic variants such as stable-predictive CFR improve the rate under predictability assumptions \citep{farina2019stable}. Excessive-gap methods give another first-order route for large sequential games \citep{kroer2018solving}.

\paragraph{Regret lower bounds in EFGs.}
Lower bounds for EFG learning depend strongly on the feedback model, the benchmark, and whether the statement concerns an algorithmic dynamic or an arbitrary learner. \citet{koolen2010hedging} proved lower bounds for structured prediction problems in terms of large orthogonal families of strategies, a principle that also applies to tree-structured decision sets. For specific game dynamics, \citet{syrgkanis2015fast} showed that multiplicative weights can suffer \(\Omega(\sqrt{T})\) regret against best responses in two-player zero-sum games, and \citet{chen2020hedging} obtained related lower bounds when both players use multiplicative weights. Complexity-theoretic barriers also arise for sparse equilibria: \citet{anagnostides2023complexity} gave oracle lower bounds for learning sparse CCEs in succinct games, including EFGs. Under bandit feedback, \citet{bai2022near} and \citet{fiegel2023adapting} established matching upper and lower bounds for learning in EFGs. Our setting is different: we study full-information reward-vector lower bounds for arbitrary online learners on treeplex strategy spaces, and the target scale is the reduced normal-form quantity \(\log|\cV|\).

\paragraph{Dilated regularization and optimality.}
Dilated regularizers were introduced by \citet{hoda2010smoothing} so that proximal steps over EFG treeplexes remain computationally tractable. Subsequent work improved the tree-parameter dependence in strong-convexity analyses of dilated entropy and dilated global entropy \citep{kroer2020faster,farina2021better}. Kernelized multiplicative weights (KOMWU) achieved the sharper \(\sqrt{T\log|\cV|}\) regret dependence on reduced normal-form size \citep{farina2022kernelized}, and \citet{bai2022efficient} showed that KOMWU is iterate-equivalent to OMD with weight-one DilEnt. The closest predecessor to this paper is \citet{fan2024optimality}, which introduced the treeplex norm pair used here, recovered the DilEnt certificate that predicts the KOMWU rate, and proved DilEnt/treeplex-norm near-optimality up to logarithmic or structural gaps. Specifically, \citet{fan2024optimality} establish a lower bound of \(\Omega(\sqrt{T\Qwidth\log|\cA_0|})\), where \(\Qwidth\) is the maximum number of terminal sequences reachable by a pure plan and \(|\cA_0|\) is the minimum action-set size. This quantity matches \(\sqrt{T\log|\cV|}\) for some restricted treeplex families, but it can leave a logarithmic gap for a general fixed treeplex. We close the remaining gap for every treeplex by reducing the comparator process to a tropical Gaussian formula and proving the exact \(\log|\cV|\) lower-bound scale using spherical entropy, Sudakov minoration \citep{vershynin2018high}, and \citet{chernozhukov2013gaussian}'s Gaussian approximation theorem for maxima.

\section{Preliminaries}
\label{sec:prelim}

\paragraph{General notation.}
For \(k\in\NN\), write \(\llbracket k\rrbracket=\{1,\ldots,k\}\). For a finite set \(\cA\), let \(\Delta^{\cA}\) be the probability simplex over \(\cA\). Vectors are bold, for example \(\xb,\yb,\wb\). If \(\xb\) is a vector and \(\cC\) is an index set, then \(\xb[\cC]\) denotes the corresponding subvector. Unless stated otherwise, \(\log\) is natural, and asymptotic notation hides only universal constants.

\paragraph{Tree-form sequential decision processes.}
We use the single-player decision space induced by an extensive-form game. Let \(\cJ\) be the set of decision points. At each \(j\in\cJ\), the learner chooses an action \(a\in\cA_j\), and the pair \(ja\) is a sequence. Define
\[
    \nonrootseqs=\{ja:j\in\cJ,\ a\in\cA_j\},
    \qquad
    \seqs=\nonrootseqs\cup\{\emptyset\}.
\]
For each decision point \(j\), let \(p_j\in\seqs\) be its parent sequence. For each nonterminal sequence \(\sigma\), let \(\cC_\sigma\subseteq\cJ\) be the set of next decision points reached after \(\sigma\). The point set is \(\tfsdp=\cJ\cup\seqs\), and \(\leaveseqs\subseteq\seqs\) is the set of terminal sequences.

The extended sequence-form strategy space is
\[
    \hat{\cQ}
    =
    \left\{\xb\in[0,1]^{\tfsdp}:
    \begin{aligned}
    &\xb[\emptyset]=1,\\
    &\xb[\sigma]=\xb[j] && \forall \sigma\in\nonleaveseqs,\ j\in\cC_\sigma,\\
    &\xb[j]=\sum_{a\in\cA_j}\xb[ja] && \forall j\in\cJ
    \end{aligned}
    \right\}.
\]
We write \(\cQ\) for the projection of \(\hat{\cQ}\) to terminal sequences. Its vertices
\[
    \cV=\cQ\cap\{0,1\}^{\leaveseqs}
\]
are the reduced normal-form pure plans. For \(h\in\tfsdp\), let \(\cV_h\) be the set of reduced normal-form pure plans restricted to the subtree rooted at \(h\), and set \(N_h\triangleq|\cV_h|\).

For a subtree \(h\), let \(\cQ_h\) be the same terminal-sequence projection restricted to terminals below \(h\), normalized so that the root of the subtree is reached with probability one. The next lemma says that every mixed sequence-form strategy on a subtree can be written as a convex combination of pure continuation plans. We use this later to argue that, when every pure plan has the same expected reward under the hard instance, every learner strategy has that same expected reward. A formal induction over the treeplex recursion is given in Appendix~\ref{app:treeplex-conv}.

\begin{lemma}[Treeplex convex hull]
\label{lem:treeplex-conv}
For every subtree \(h\), \(\cQ_h=\co(\cV_h)\). In particular, \(\cQ=\co(\cV)\).
\end{lemma}

\paragraph{Transition kernels and reward vectors.}
We next describe the reward vectors used in the online-learning problem. A transition kernel is represented by \(\yb\in[0,1]^{\tfsdp}\) satisfying
\[
    \yb[\emptyset]=1,\qquad
    \yb[\sigma]=\sum_{j\in\cC_\sigma}\yb[j]\quad \forall \sigma\in\nonleaveseqs,
    \qquad
    \yb[j]=\yb[ja]\quad \forall j\in\cJ,\ a\in\cA_j.
\]
Let \(\cY\) be the set of such kernels. The coordinate \(\yb[\sigma]\) is the reach multiplier contributed by chance and the non-learner players at terminal sequence \(\sigma\). Given signed terminal rewards \(\rb\in[-1,1]^{\leaveseqs}\), define the full-information reward vector by
\[
    \wb(\yb,\rb)[\sigma]\triangleq \yb[\sigma]\rb[\sigma],
    \qquad \sigma\in\leaveseqs.
\]
Thus a strategy \(\xb\in\cQ\) receives reward \(\langle\xb,\wb(\yb,\rb)\rangle\). This convention folds transition reach into the linear reward vector, while \(\rb\) remains the terminal reward conditional on reaching the terminal.

\paragraph{Online learning and regret.}
At each round \(t\), the learner selects \(\xb_t\in\cQ\) and then observes a reward vector \(\wb_t\in\RR^{\leaveseqs}\). The cumulative external regret is
\[
    \Regret(T)
    \triangleq
    \max_{\xb_*\in\cQ}\sum_{t=1}^T
    \langle \xb_*-\xb_t,\wb_t\rangle.
\]
We focus on full-information rewards induced by bounded terminal rewards and valid transition kernels. The hard instances below use Rademacher terminal rewards in \(\{-1,1\}\).

\paragraph{Treeplex norms and DilEnt.}
Following \citet{fan2024optimality}, define the treeplex norm pair
\[
    \|\ub\|\Ynorm \triangleq \sup_{\yb\in\cY}\langle |\ub|,\yb\rangle,
    \qquad
    \|\ub\|\Qnorm \triangleq \sup_{\xb\in\cQ}\langle |\ub|,\xb\rangle,
\]
where \(\cY\) is the transition-kernel space. When the treeplex is a simplex, these reduce to the usual \(\ell_1/\ell_\infty\) pair. The weight-one dilated entropy regularizer is
\[
    \DilEnt(\xb)
    \triangleq
    \sum_{j\in\cJ}\sum_{a\in\cA_j}
    \xb[ja]\log\frac{\xb[ja]}{\xb[p_j]},
\]
with the convention that \(0\log 0=0\). The facts we need are the following: DilEnt is \(1\)-strongly convex with respect to \(\|\cdot\|\Ynorm\), every feasible reward vector satisfies \(\|\wb_t\|\Qnorm\le 1\), and the DilEnt Bregman diameter from its minimizer is at most \(\log|\cV|\). Therefore, OMD with DilEnt gives
\[
    \Regret(T)\le \cO(\sqrt{T\log|\cV|}).
\]
The lower bound below shows that no online learner can improve this logarithmic dependence.

Our hard instance uses signed rewards in \(\{-1,1\}\), which match the Rademacher notation and are equivalent to the usual \([0,1]\) normalization up to constants.

\section{Tropical Gaussian Formulae}
\label{sec:formula}

For \(K\) independent experts, the classical lower bound follows from the maximum of \(K\) independent Gaussian or Rademacher sums. Reduced normal-form plans are not independent experts, because different plans reuse decisions and share terminal outcomes. To preserve these dependencies, we introduce tropical Gaussian formulae, the recursive analogue of this construction. We now define the Gaussian formula used in the lower bound. The definition follows the same recursion as pure plans in a treeplex. At decision points, a pure plan chooses one continuation, so the formula uses a maximum. At observation points, a pure plan must carry continuations for all successor decision points, so the formula combines independent subformulae by a normalized sum. The number \(N_Z\) counts the indices, or equivalently the pure continuations, of the formula.

\begin{definition}[Tropical Gaussian formula]
A tropical Gaussian formula is generated recursively by
\begin{itemize}[leftmargin=*]
    \item A leaf is a standard Gaussian $G\sim\cN(0,1)$ and has index count $N_G=1$.
    \item If $X,Y$ are formulae, then $Z=\max\{X,Y\}$ is a formula with $N_Z=N_X+N_Y$.
    \item If $X,Y$ are formulae and $a\in[0,1]$, then \(Z=\sqrt a\,X+\sqrt{1-a}\,Y\) is a formula with $N_Z=N_XN_Y$.
\end{itemize}
Formulae are rooted trees. Each occurrence of a subformula uses fresh independent Gaussian leaves, so disjoint subtrees are independent before they are combined. We call the third operation a mix node. The formula is \emph{log-balanced} if every mix node uses
\[
    a=\frac{\log N_X}{\log N_X+\log N_Y},
\]
with the convention that when \(N_X=N_Y=1\), any \(a\in[0,1]\) may be used.
\end{definition}

The word ``tropical'' refers to the max operation, following the convention from tropical algebra where ordinary addition is replaced by max or min~\citep{richter2005first}. The word ``Gaussian'' refers to the independent Gaussian leaves and to the normalized sums, whose squared coefficients add to one. Max nodes represent choices among alternatives. Mix nodes combine independent subproblems. This matches the treeplex structure: a pure plan chooses one action at a decision point, but specifies continuations at every successor information set after an observation point.

The next lemma rewrites every formula as the maximum of finitely many Gaussian linear forms. After this step, lower bounding \(\Expt[Z]\) becomes a question about the geometry of \(N_Z\) unit vectors: if many of these vectors are well separated, then the expected maximum is large. The recursive construction of the vectors is given in Appendix~\ref{app:formula-proof}.

\begin{lemma}[Gaussian-process representation]
\label{lem:formula-gp-representation}
Every tropical Gaussian formula \(Z\) admits an index set \(\cI_Z\), a dimension \(d_Z\), and a map \(\Phi_Z:\cI_Z\to\mathbb S^{d_Z-1}\) such that, for \(\gb_Z\sim\cN(0,I_{d_Z})\),
\[
    Z\stackrel{d}{=}
    \max_{\iota\in\cI_Z}\langle \gb_Z,\Phi_Z(\iota)\rangle,
    \qquad
    |\cI_Z|=N_Z.
\]
\end{lemma}

The index count \(N_Z\) tells us how many Gaussian linear forms appear in the representation, but it does not tell us how different these linear forms are. The vectors \(\Phi_Z(\iota)\) need not be orthogonal or even well separated. Thus, to lower bound \(\Expt[Z]\), we need to measure how many of these vectors remain distinguishable at each geometric scale.

\begin{definition}[Spherical entropy profile]
\label{def:spherical-entropy}
Fix the representation from Lemma~\ref{lem:formula-gp-representation}. For independent \(U,V\sim\Unif(\cI_Z)\), define the overlap distance
\[
    D_Z(U,V)
    \triangleq
    \frac12\|\Phi_Z(U)-\Phi_Z(V)\|_2^2
    =
    1-\langle\Phi_Z(U),\Phi_Z(V)\rangle.
\]
We measure separation through the average close-pair weight
\[
    A_Z(\lambda)
    \triangleq
    \Expt_{U,V}\exp\bigl(-\lambda D_Z(U,V)\bigr),
\]
and define the spherical entropy profile by
\[
    B_Z(\lambda)\triangleq-\log A_Z(\lambda).
\]
\end{definition}

The profile \(B_Z\) should be read as an effective version of \(\log N_Z\) at a given scale. The raw count \(N_Z\) only counts how many indices appear in the Gaussian-process representation; it does not say whether the corresponding vectors are actually separated. In contrast, \(A_Z(\lambda)\) is large when two random indices often produce nearby vectors, and \(B_Z(\lambda)=-\log A_Z(\lambda)\) is large when such close pairs are rare. Thus \(B_Z(\lambda)\) measures how many directions remain distinguishable at resolution \(\lambda^{-1/2}\).

The next lemma gives the exact recursion for this close-pair quantity under the two formula operations. Its derivation from the Gaussian-process representation is given in Appendix~\ref{app:formula-proof}.

\begin{lemma}[Close-pair recursion]
\label{lem:overlap-transform-recursion}
The profile in Definition~\ref{def:spherical-entropy} obeys the following exact recursions. For a leaf, \(A_G(\lambda)=1\). If \(Z=\max\{X,Y\}\), \(n=N_X\), \(m=N_Y\), and \(N=n+m\), then
\[
    A_Z(\lambda)
    =
    \biggl(\frac nN\biggr)^2 A_X(\lambda)
    +
    \biggl(\frac mN\biggr)^2 A_Y(\lambda)
    +
    \frac{2nm}{N^2}e^{-\lambda}.
\]
For a mix node \(Z=\sqrt a\,X+\sqrt{1-a}\,Y\), the recursion is
\[
    A_Z(\lambda)=A_X(a\lambda)A_Y((1-a)\lambda).
\]
\end{lemma}

We use this profile in two steps. First, a large value of \(B_Z(\lambda)\) at any scale \(\lambda\) implies that many Gaussian directions are separated, which gives a lower bound on \(\Expt[Z]\). Second, for log-balanced formulae, the recursion above shows that \(B_Z(\lambda)\) is large at the scale \(\lambda=2\log N_Z\). We state the first, scale-free step separately because it is also useful for non-log-balanced routed laws.

\begin{theorem}[Entropy-to-width lower bound]
\label{thm:effective-entropy}
For every tropical Gaussian formula \(Z\), there exists a universal constant \(c>0\) such that
\[
    \Expt[Z]
    \ge
    c
    \sup_{\lambda>0:\,B_Z(\lambda)\ge2}
    \frac{B_Z(\lambda)}{\sqrt{\lambda}}.
\]
Define the effective entropy by
\[
    \Heff(Z)
    \triangleq
    \sup_{\lambda>0:\,B_Z(\lambda)\ge2}
    \frac{B_Z(\lambda)^2}{\lambda},
\]
so \(\Expt[Z]\ge c\sqrt{\Heff(Z)}\). If no \(\lambda\) satisfies \(B_Z(\lambda)\ge2\), the supremum is defined to be zero.
\end{theorem}

\begin{proof}[Proof sketch]
Fix a scale \(\lambda\) with \(B=B_Z(\lambda)\ge2\). Since \(A_Z(\lambda)=e^{-B}\), Markov's inequality implies that only an \(e^{-B/2}\) fraction of index pairs $(\iota, \kappa)$ can satisfy
\[
    \|\Phi_Z(\iota)-\Phi_Z(\kappa)\|_2
    \le
    \sqrt{\frac{B}{\lambda}}.
\]
Build a graph on \(\cI_Z\) by connecting such close pairs. Tur\'an's theorem gives an independent set \(\cJ_Z\) of size at least 
\[
\begin{aligned}
    |\cJ_Z|
    &\ge
    \frac{N_Z^2}{2|E(G_{\mathrm{close}})|+N_Z}
    \ge c_1e^{B/2}, \qquad s.t. \quad
    \|\Phi_Z(\iota)-\Phi_Z(\kappa)\|_2
    \ge
    \sqrt{\frac{B}{\lambda}}
    \qquad
    \forall \iota\ne\kappa\in\cJ_Z.
\end{aligned}
\]
Sudakov minoration then yields
\[
    \Expt[Z]
    \ge
    c_2\sqrt{\frac{B}{\lambda}}\sqrt{\log|\cJ_Z|}
    \ge
    c\frac{B}{\sqrt\lambda}.
\]
Taking the supremum over all scales with \(B_Z(\lambda)\ge2\) proves the theorem. The full graph-counting argument is given in Appendix~\ref{app:effective-entropy}.
\end{proof}

We now specialize to log-balanced formulae. In this case, the mix weights are chosen according to logarithmic subtree sizes, and the recursion gives a simple estimate at the scale \(\lambda=2\log N_Z\).

\begin{lemma}[Log-balanced entropy scale]
\label{lem:logbalanced-transform}
Suppose every mix node is log-balanced. Then every tropical Gaussian formula \(Z\) satisfies
\[
    A_Z(2\log N_Z)\le N_Z^{-1/2}.
\]
\end{lemma}

Since \(B_Z(\lambda)=-\log A_Z(\lambda)\), the lemma gives \(B_Z(2\log N_Z)\ge (1/2) \cdot \log N_Z\). Thus one scale in Theorem~\ref{thm:effective-entropy} already contributes order \(\sqrt{\log N_Z}\).

\begin{theorem}[Formula lower bound]
\label{thm:formula-lb}
There exists a universal constant \(c>0\) such that every log-balanced tropical Gaussian formula \(Z\) satisfies
\[
    \Expt[Z]\ge c\sqrt{\log N_Z}.
\]
\end{theorem}

\begin{proof}[Proof sketch]
Set \(\lambda=2\log N_Z\). By Lemma~\ref{lem:logbalanced-transform}, \(B_Z(\lambda)\ge\frac12\log N_Z\). For all sufficiently large \(N_Z\), Theorem~\ref{thm:effective-entropy} applies and gives
\[
    \Expt[Z]
    \ge
    c_0\frac{B_Z(\lambda)}{\sqrt{\lambda}}
    \ge
    \frac{c_0}{2\sqrt2}\sqrt{\log N_Z}.
\]
After adjusting the universal constant and handling the bounded small-\(N_Z\) cases, this proves the theorem. The full induction and constants are given in Appendix~\ref{app:formula-proof}.
\end{proof}

The main reduction below uses only the log-balanced consequence in Theorem~\ref{thm:formula-lb}. The general entropy-to-width theorem is used again in Appendix~\ref{app:arbitrary-routing}, where non-log-balanced routing can leave fewer directions separated at the relevant scales.
\section{The Treeplex Hard Instance}
\label{sec:reduction}

We now turn the formula lower bound into an online-learning lower bound. The first step is to match the recursion of pure plans with the recursion of tropical Gaussian formulae. At a decision point, a pure plan chooses one action, so continuation counts add over actions. At an observation point, a pure plan specifies continuations for all successor information sets, so continuation counts multiply across those successor decision points.

\begin{lemma}[Subtree plan-count recursion]
\label{lem:subtree-plan-counts}
For every point \(h\in\tfsdp\), let \(\cV_h\) be the set of reduced normal-form plans restricted to the subtree rooted at \(h\), let \(\leaveseqs_h\) be the terminals below \(h\), and let \(N_h=|\cV_h|\). The count satisfies \(N_\sigma=1\) for every \(\sigma\in\leaveseqs\), and the recursion is
\[
    N_j=\sum_{a\in\cA_j} N_{ja},
    \qquad
    N_\sigma=\prod_{j\in\cC_\sigma}N_j
    \quad \text{for } \sigma\in\nonleaveseqs.
\]
\end{lemma}

This add-multiply recursion is exactly the recursion of the Gaussian formula: decision points become max nodes, and observation points become mix nodes. The hard instance realizes this structure using signed Rademacher terminal rewards. At a decision point, the child one-round processes in all action subtrees are sampled independently, and a pure plan selects the subtree corresponding to its chosen action. At an observation point, the adversary samples one successor \(j_r\) with probability proportional to \(\log N_{j_r}\), routes all transition mass to that successor, and uses the child one-round process there. The detailed recursive construction is given in Appendix~\ref{app:treeplex-reduction}; the next proposition states the invariant needed for the lower bound.

\begin{proposition}[Treeplex-to-formula reduction]
\label{prop:treeplex-formula-reduction}
For every treeplex, there exists a log-balanced tropical Gaussian formula \(Z_\emptyset\) and an oblivious one-round distribution over feasible reward vectors \(\wb=\wb(\yb,\rb)\), with valid \(\yb\in\cY\) and \(\rb\in[-1,1]^{\leaveseqs}\), such that
\[
    N_{Z_\emptyset}=|\cV|.
\]
Moreover, for every pure plan \(\pi\in\cV\), the reward \(W(\pi)=\langle\vb_\pi,\wb\rangle\) has mean \(0\), and the process \((W(\pi))_{\pi\in\cV}\) has the same covariance kernel as the Gaussian-process representation of \(Z_\emptyset\). Consequently, if \(G_\emptyset\) is the centered Gaussian process with this covariance, then
\[
    Z_\emptyset\stackrel{d}{=}\max_{\pi\in\cV}G_\emptyset(\pi).
\]
\end{proposition}

\begin{proof}[Proof sketch]
The plan-count identity is Lemma~\ref{lem:subtree-plan-counts}. Feasibility follows because routing all transition mass to one successor defines a valid transition kernel; terminal rewards outside the routed branch have zero reach and therefore do not affect any realized payoff. The covariance induction matches the formula operations: independent action subtrees give orthogonal covariance blocks, while random routing gives the weighted covariance average of a mix node.
\end{proof}
We now sample this one-round law independently for \(T\) rounds. Since each pure plan has zero one-round mean, any learner also has conditional expected reward \(0\) at each round. Thus the regret is driven by the best fixed pure plan, whose cumulative signed gain is
\[
    M_T
    \triangleq
    \max_{\pi\in\cV}
    \sum_{t=1}^T W_t(\pi).
\]
The finite-time comparison in Appendix~\ref{app:rademacher-gaussian-comparison} relates \(M_T/\sqrt T\) to the Gaussian maximum \(Z_\emptyset\). Combining this comparison with the formula lower bound gives the regret lower bound.

\begin{theorem}[Finite-time EFG regret lower bound]
\label{thm:efg-lb}
There exist universal constants \(c,C>0\) such that, for every treeplex with reduced normal-form strategy set \(\cV\) and every \(T\ge C\log^{14}(|\cV|+1)\), there exists an oblivious distribution over terminal reward vectors \(\rb_t\in[-1,1]^{\leaveseqs}\) and valid transition kernels \(\yb_t\in\cY\), obtained by independently sampling routed successors with log-balanced probabilities at observation points, such that the feasible reward vectors \(\wb_t=\wb(\yb_t,\rb_t)\) make every online learning algorithm satisfy
\[
    \Expt[\Regret(T)]
    \ge
    c\sqrt{T\log|\cV|}.
\]
\end{theorem}

\begin{proof}[Proof sketch]
Run the reduction at the root for \(T\) independent rounds. By Proposition~\ref{prop:treeplex-formula-reduction}, every pure plan has mean-zero reward in each round. By linearity, every learner's expected cumulative reward is therefore \(0\). The best fixed pure plan obtains \(M_T\). By the finite-time comparison, \(\Expt[M_T/\sqrt T]\) is at least \(\Expt[Z_\emptyset]\) up to a lower-order error. By Theorem~\ref{thm:formula-lb}, \(\Expt[Z_\emptyset]\ge c\sqrt{\log|\cV|}\). Since \(T\ge C\log^{14}(|\cV|+1)\), the comparison error is lower order, so \(\Expt[\Regret(T)]\ge c\sqrt{T\log|\cV|}\). The full argument is in Appendix~\ref{app:efg-lower-bound}.
\end{proof}

The log-balanced routing is what makes the Gaussian width at the root scale as \(\sqrt{\log|\cV|}\). If the routed successors are sampled with other fixed probabilities, the same reduction gives the effective-entropy variant in Appendix~\ref{app:arbitrary-routing}.

\section{Minimax Optimality of DilEnt}
\label{sec:optimality}

The lower bound also shows that the logarithmic scale achieved by weight-one DilEnt is unavoidable for the standard OMD analysis. Consider any distance-generating function \(\varphi\) used with OMD on the treeplex strategy space \(\cQ\). Suppose \(\varphi\) is \(\mu\)-strongly convex with respect to a norm \(\|\cdot\|\), and suppose every feasible reward vector satisfies \(\|\wb\|_*\le1\). Let
\[
    D_\varphi
    \triangleq
    \sup_{\xb_*\in\cQ}
    \cD_\varphi(\xb_*\|\xb_1),
    \qquad
    \xb_1\in\argmin_{\xb\in\cQ}\varphi(\xb).
\]
The standard OMD guarantee gives
\[
    \Regret(T)
    \le
    \frac{D_\varphi}{\eta}+\frac{\eta T}{2\mu}.
\]
Optimizing \(\eta\) yields
\[
    \Regret(T)
    \le
    \cO\Bigl(\sqrt{D_\varphi T/\mu}\Bigr).
\]

Thus the ratio \(D_\varphi/\mu\) is the quantity that determines the regret scale in this proof template. If this ratio were smaller than order \(\log|\cV|\), while the dual norm still controlled all feasible reward vectors, the resulting OMD guarantee would contradict the lower bound from Theorem~\ref{thm:efg-lb}.

\begin{corollary}[Optimal OMD scale]
\label{cor:dil-ent-optimal}
Any regularizer-and-norm pair that gives the above OMD guarantee uniformly over feasible EFG reward vectors must satisfy
\[
    \frac{D_\varphi}{\mu}\ge \Omega(\log|\cV|).
\]
Weight-one DilEnt satisfies \(D_{\DilEnt}/\mu\le \log|\cV|\) under the treeplex norm pair \(\|\cdot\|\Ynorm,\|\cdot\|\Qnorm\). Hence DilEnt has the optimal diameter-to-strong-convexity scale among such OMD guarantees.
\end{corollary}

\begin{proof}[Proof sketch]
The OMD guarantee applies to the hard instance from Theorem~\ref{thm:efg-lb} whenever all feasible reward vectors satisfy \(\|\wb_t\|_*\le1\). If \(D_\varphi/\mu=o(\log|\cV|)\), then the optimized OMD bound would be \(o(\sqrt{T\log|\cV|})\), contradicting Theorem~\ref{thm:efg-lb} for sufficiently large \(T\). Therefore \(D_\varphi/\mu=\Omega(\log|\cV|)\). Weight-one DilEnt matches this scale by the treeplex-norm diameter and strong-convexity facts stated in Section~\ref{sec:prelim}. The formal proof is given in Appendix~\ref{app:dilent-optimality}.
\end{proof}

The ratio \(D_\varphi/\mu\) is invariant under rescaling \(\varphi\), since both \(D_\varphi\) and \(\mu\) scale by the same factor. A different norm can improve the OMD bound only if its dual norm still controls every feasible reward vector; otherwise the guarantee does not apply to the hard instance. DilEnt is therefore optimal within the standard OMD proof template: improving the \(\log|\cV|\) scale would require giving up uniform control of feasible reward vectors.

Combined with the upper bounds of \citet{farina2022kernelized} and \citet{bai2022efficient}, Theorem~\ref{thm:efg-lb} resolves the question left open by \citet{fan2024optimality}: weight-one DilEnt and KOMWU have the minimax-optimal \(\log|\cV|\) dependence on every treeplex, and the logarithm of the reduced normal-form strategy count is the intrinsic complexity of the problem.

\section{Conclusion}

We proved a minimax lower bound for full-information online learning on treeplexes by reducing the problem to a tropical Gaussian formula lower bound. The hard instance uses Rademacher terminal rewards; the Gaussian process is only an analytic tool that captures the covariance among reduced normal-form pure plans. The lower bound shows that the \(\log|\cV|\) dependence is not an artifact of KOMWU or of a loose upper bound, but is forced by the recursive decision-and-observation structure of the treeplex. Thus any standard OMD analysis whose dual norm controls all feasible reward vectors must have diameter-to-strong-convexity ratio \(\Omega(\log|\cV|)\), and weight-one DilEnt matches that scale through the treeplex-norm certificate.

\begin{ack}
This work was supported in part by National Science Foundation awards CCF-2443068 and IIS-2552046, Office of Naval Research grants N000142512296 and N00014-24-1-2470, and an AI2050 Early Career Fellowship. Part of this research was performed while A.S. was visiting the Institute for Mathematical and Statistical Innovation (IMSI), which is supported by the National Science Foundation (Grant No. DMS-2425650).
\end{ack}

\bibliographystyle{plainnat}
\bibliography{refs}

\clearpage
\appendix

{\LARGE Appendix}

\section{Treeplex Convex Hull}
\label{app:treeplex-conv}

\begin{proof}[Proof of Lemma~\ref{lem:treeplex-conv}]
We prove the stronger subtree statement by induction from the leaves upward. Write \(\cQ_h\) for terminal vectors below \(h\), normalized to have reach one at \(h\).

If \(h=\sigma\in\leaveseqs\), then \(\cQ_\sigma=\cV_\sigma=\{1\}\). Suppose \(h=j\) is a decision point. Any \(\xb\in\cQ_j\) is determined by action masses \(\alpha_a=\xb[ja]\), with \(\sum_a\alpha_a=1\), and by normalized child vectors \(\xb_{ja}/\alpha_a\in\cQ_{ja}\) whenever \(\alpha_a>0\). By induction, each child vector is a convex combination of \(\cV_{ja}\). Therefore \(\xb\) is a convex combination of pure plans that choose one action \(a\) at \(j\) and then use one pure child continuation in \(\cV_{ja}\). These are exactly \(\cV_j\).

Now suppose \(h=\sigma\) is a nonterminal observation point with successors \(\cC_\sigma=\{j_1,\ldots,j_m\}\). A feasible subtree strategy is a tuple \((\xb_{j_1},\ldots,\xb_{j_m})\) with \(\xb_{j_r}\in\cQ_{j_r}\). By induction, write
\[
    \xb_{j_r}=\sum_{\nu_r\in\cV_{j_r}}\alpha_{r,\nu_r}\vb_{\nu_r}.
\]
Sampling the child pure plans independently with probabilities \(\alpha_{r,\nu_r}\) gives
\[
    (\xb_{j_1},\ldots,\xb_{j_m})
    =
    \sum_{\nu_1,\ldots,\nu_m}
    \biggl(\prod_{r=1}^m\alpha_{r,\nu_r}\biggr)
    (\vb_{\nu_1},\ldots,\vb_{\nu_m}).
\]
The tuple pure plans on the right are precisely the reduced pure plans in \(\cV_\sigma\). Hence \(\cQ_\sigma\subseteq\co(\cV_\sigma)\). The reverse inclusion is immediate because every pure plan is feasible and \(\cQ_h\) is convex. Taking \(h=\emptyset\) gives \(\cQ=\co(\cV)\).
\end{proof}

\section{Auxiliary Packing and Gaussian-Width Inequalities}
\label{app:packing-width-tools}

We present two standard tools on formula lower bound. The first turns a sparse close-pair graph into a large separated subset. The second turns that separation into a lower bound on Gaussian width.

Following the idea of Tur\'an's theorem (see, e.g., \citet{bollobas1998modern}), we prove the following close-pair extraction lemma, which gives a lower bound on the size of an independent set.

\begin{lemma}[Close-pair extraction]
\label{lem:close-pair-extraction}
Let \(S\) be a finite set with \(|S|=N\ge1\). Let \(R\subseteq S\times S\)
be a symmetric relation and suppose that
\[
    |R|\le \eta N^2
\]
for some \(\eta\ge0\). Build a simple graph \(G_{\mathrm{close}}\) on vertex
set \(S\) by connecting distinct \(x,y\in S\) whenever \((x,y)\in R\). Then
there exists an independent set \(J\subseteq S\) such that
\[
    |J|
    \ge
    \frac{N^2}{\eta N^2+N}
    =
    \frac{N}{1+\eta N}.
\]
In particular, no two distinct points of \(J\) are adjacent in
\(G_{\mathrm{close}}\).
\end{lemma}

\begin{proof}
Let \(E_{\mathrm{close}}\) be the edge set of \(G_{\mathrm{close}}\), and
write \(M=|E_{\mathrm{close}}|\). Since \(R\) is symmetric, each edge
\(\{x,y\}\in E_{\mathrm{close}}\) contributes the two ordered pairs
\((x,y)\) and \((y,x)\) to \(R\). Hence
\[
    2M\le |R|\le \eta N^2 .
\]

We now prove directly that \(G_{\mathrm{close}}\) has an independent set of
size at least \(N^2/(N+2M)\). Choose a uniformly random ordering of the
vertices of \(S\). Let \(J\) be the set of vertices that appear before all of
their neighbors in this ordering.

The set \(J\) is independent: if two adjacent vertices \(x\) and \(y\) both
belonged to \(J\), then \(x\) would have to appear before \(y\), since \(y\)
is a neighbor of \(x\), and \(y\) would have to appear before \(x\), since
\(x\) is a neighbor of \(y\), a contradiction.

For each \(x\in S\), let \(d(x)\) be its degree in \(G_{\mathrm{close}}\).
Among the \(d(x)+1\) vertices consisting of \(x\) and its neighbors, each is
equally likely to appear first in the random ordering. Therefore
\[
    \mathbb{P}(x\in J)=\frac{1}{d(x)+1}.
\]
By linearity of expectation,
\[
    \mathbb{E}|J|
    =
    \sum_{x\in S}\frac{1}{d(x)+1}.
\]
Using Cauchy's inequality,
\[
    \sum_{x\in S}\frac{1}{d(x)+1}
    \ge
    \frac{N^2}{\sum_{x\in S}(d(x)+1)}
    =
    \frac{N^2}{N+2M}.
\]
Hence there exists at least one ordering for which the corresponding
independent set \(J\) satisfies
\[
    |J|
    \ge
    \frac{N^2}{N+2M}.
\]
Finally, since \(2M\le \eta N^2\), we obtain
\[
    |J|
    \ge
    \frac{N^2}{N+\eta N^2}
    =
    \frac{N}{1+\eta N}.
\]
This proves the claim.
\end{proof}

The second tool is Sudakov minoration for Gaussian processes. We use the
following finite separated form, which follows from the standard entropy form
of Sudakov minoration; see \citep{vershynin2018high}.

\begin{theorem}[Sudakov minoration, finite separated form]
\label{thm:sudakov-minoration-finite}
Let \(X_1,\ldots,X_M\) be a centered jointly Gaussian family, with \(M\ge2\).
Suppose that there exists \(\Delta>0\) such that
\[
    \bigl(\Expt[(X_i-X_j)^2]\bigr)^{1/2}
    \ge
    \Delta
    \qquad
    \text{for all } i\ne j .
\]
Then there exists a universal constant \(c>0\) such that
\[
    \Expt\max_{1\le i\le M} X_i
    \ge
    c\,\Delta\,\sqrt{\log M}.
\]
\end{theorem}

\begin{proof}[Derivation from the entropy form]
Let
\[
    d(i,j)=\bigl(\Expt[(X_i-X_j)^2]\bigr)^{1/2}
\]
be the canonical metric. Since the points \(1,\ldots,M\) are
\(\Delta\)-separated in \(d\), every \(d\)-ball of radius \(\Delta/4\)
contains at most one point. Therefore the covering number satisfies
\[
    \mathcal N(\llbracket M\rrbracket,d,\Delta/4)\ge M .
\]
The standard entropy form of Sudakov minoration gives
\[
    \Expt\max_{1\le i\le M}X_i
    =
    \Expt\sup_{i\in\llbracket M\rrbracket}X_i
    \ge
    c_0\frac{\Delta}{4}
    \sqrt{
        \log \mathcal N(\llbracket M\rrbracket,d,\Delta/4)
    }
    \ge
    c\,\Delta\sqrt{\log M},
\]
after absorbing the factor \(1/4\) into the universal constant.
\end{proof}

For the Gaussian-process representation of tropical Gaussian formulae, we use
the following Euclidean consequence.

\begin{lemma}[Sudakov minoration for separated Gaussian linear forms]
\label{lem:sudakov-width-separated}
Let \(T=\{v_1,\ldots,v_M\}\subseteq\RR^d\), let \(\gb\sim\cN(0,I_d)\), and
suppose that \(M\ge2\) and
\[
    \|v_i-v_j\|_2\ge \Delta
    \qquad
    \text{for all } i\ne j .
\]
Then there exists a universal constant \(c>0\) such that
\[
    \Expt\max_{1\le i\le M}\langle \gb,v_i\rangle
    \ge
    c\,\Delta\,\sqrt{\log M}.
\]
\end{lemma}

\begin{proof}
Set \(X_i=\langle \gb,v_i\rangle\). Then \(X_1,\ldots,X_M\) are centered
jointly Gaussian and
\[
    \bigl(\Expt[(X_i-X_j)^2]\bigr)^{1/2}
    =
    \|v_i-v_j\|_2 .
\]
The result follows from Theorem~\ref{thm:sudakov-minoration-finite}.
\end{proof}

\section{Proof of the Formula Lower Bound}
\label{app:formula-proof}

\begin{proof}[Proof of Lemma~\ref{lem:formula-gp-representation}]
We construct the representation by induction over the formula tree. For a leaf \(G\), set \(\cI_G=\{\star\}\), \(d_G=1\), and \(\Phi_G(\star)=1\). Then, for \(g_G\sim\cN(0,1)\),
\[
    \max_{\iota\in\cI_G}\langle g_G,\Phi_G(\iota)\rangle
    =
    g_G
    \stackrel{d}{=}G,
\]
and \(|\cI_G|=1=N_G\).

Suppose next that \(Z=\max\{X,Y\}\), and assume the claim for \(X\) and \(Y\). Let \(d_Z=d_X+d_Y\) and set
\[
    \cI_Z=(\cI_X\times\{0\})\cup(\cI_Y\times\{1\}),
\]
Define the embedded maps by
\[
    \Phi_Z(\iota,0)=(\Phi_X(\iota),0),
    \qquad
    \Phi_Z(\kappa,1)=(0,\Phi_Y(\kappa)).
\]
Each image vector has unit norm. For \(\gb_Z=(\gb_X,\gb_Y)\) with independent standard Gaussian blocks,
\[
    \max_{\zeta\in\cI_Z}\langle \gb_Z,\Phi_Z(\zeta)\rangle
    =
    \max\biggl\{
    \max_{\iota\in\cI_X}\langle\gb_X,\Phi_X(\iota)\rangle,
    \max_{\kappa\in\cI_Y}\langle\gb_Y,\Phi_Y(\kappa)\rangle
    \biggr\}
    \stackrel{d}{=}\max\{X,Y\}.
\]
Moreover, \(|\cI_Z|=|\cI_X|+|\cI_Y|=N_X+N_Y=N_Z\).

Finally suppose \(Z=\sqrt aX+\sqrt{1-a}Y\), with \(a\in[0,1]\), and again assume the claim for \(X\) and \(Y\). Let \(d_Z=d_X+d_Y\) and set \(\cI_Z=\cI_X\times\cI_Y\). Define
\[
    \Phi_Z(\iota,\kappa)
    =
    \bigl(\sqrt a\,\Phi_X(\iota),
          \sqrt{1-a}\,\Phi_Y(\kappa)\bigr).
\]
These vectors are again unit vectors. For independent Gaussian blocks \(\gb_X,\gb_Y\),
\[
    \max_{(\iota,\kappa)\in\cI_Z}
    \langle \gb_Z,\Phi_Z(\iota,\kappa)\rangle
    =
    \sqrt a
    \max_{\iota\in\cI_X}\langle\gb_X,\Phi_X(\iota)\rangle
    +
    \sqrt{1-a}
    \max_{\kappa\in\cI_Y}\langle\gb_Y,\Phi_Y(\kappa)\rangle.
\]
Because \(\cI_Z=\cI_X\times\cI_Y\), the maximum separates over the two coordinates. The right-hand side has the same distribution as \(\sqrt aX+\sqrt{1-a}Y\), and \(|\cI_Z|=|\cI_X||\cI_Y|=N_XN_Y=N_Z\).
\end{proof}

For two independent uniformly sampled indices \(U,V\sim\Unif(\cI_Z)\), recall that
\[
    D_Z(U,V)
    \triangleq
    \frac12\|\Phi_Z(U)-\Phi_Z(V)\|_2^2
    =
    1-\langle\Phi_Z(U),\Phi_Z(V)\rangle.
\]

\begin{proof}[Proof of Lemma~\ref{lem:overlap-transform-recursion}]
For a leaf, \(\cI_G\) has one element and \(D_G(U,V)=0\), hence \(A_G(\lambda)=1\).

Suppose \(Z=\max\{X,Y\}\). Write \(n=N_X\), \(m=N_Y\), and \(N=n+m\). A uniformly sampled index of \(\cI_Z\) lies in the \(X\)-block with probability \(n/N\) and in the \(Y\)-block with probability \(m/N\). If both sampled indices lie in the \(X\)-block, the distance is \(D_X\); if both lie in the \(Y\)-block, the distance is \(D_Y\). In the cross case, the two representation vectors lie in orthogonal coordinate subspaces, so their inner product is zero and \(D_Z=1\). Therefore
\[
    A_Z(\lambda)
    =
    \biggl(\frac nN\biggr)^2 A_X(\lambda)
    +
    \biggl(\frac mN\biggr)^2 A_Y(\lambda)
    +
    \frac{2nm}{N^2}e^{-\lambda}.
\]

Finally suppose \(Z=\sqrt aX+\sqrt{1-a}Y\). Let \(U=(U_X,U_Y)\) and \(V=(V_X,V_Y)\). Coordinatewise additivity gives the distance formula
\[
    D_Z(U,V)=aD_X(U_X,V_X)+(1-a)D_Y(U_Y,V_Y).
\]
The pairs \((U_X,V_X)\) and \((U_Y,V_Y)\) are independent, so
\[
\begin{aligned}
    A_Z(\lambda)
    &=
    \Expt\exp\bigl(-a\lambda D_X-(1-a)\lambda D_Y\bigr)\\
    &=
    A_X(a\lambda)A_Y((1-a)\lambda).
\end{aligned}
\]
This proves the exact recursions.
\end{proof}

\begin{proof}[Proof of Lemma~\ref{lem:logbalanced-transform}]
We prove the statement by induction over the formula tree. The transform \(A_Z(\lambda)\) is nonincreasing in \(\lambda\), because \(D_Z\ge0\). For a leaf, \(N_Z=1\), \(D_Z(U,V)=0\), and \(A_Z(2\log N_Z)=A_Z(0)=1=N_Z^{-1/2}\).

Suppose first that $Z=\sqrt aX+\sqrt{1-a}Y$, with
\[
    a=\frac{\log N_X}{\log N_X+\log N_Y}.
\]
Let $L_X=\log N_X$, $L_Y=\log N_Y$, and $L=L_X+L_Y=\log N_Z$. If \(L=0\), then \(N_Z=1\) and the claim is trivial. Otherwise, Lemma~\ref{lem:overlap-transform-recursion} and the log-balanced choice give
\[
\begin{aligned}
    A_Z(2L)
    &=
    A_X(2L_X)A_Y(2L_Y)\\
    &\le
    N_X^{-1/2}N_Y^{-1/2}
    =
    N_Z^{-1/2}.
\end{aligned}
\]

Now suppose $Z=\max\{X,Y\}$. Let $n=N_X$, $m=N_Y$, $N=N_Z=n+m$, and \(L=\log N\). Since \(L\ge\log n\) and \(L\ge\log m\), monotonicity and the induction hypothesis give
\[
    A_X(2L)\le n^{-1/2},
    \qquad
    A_Y(2L)\le m^{-1/2}.
\]
Applying Lemma~\ref{lem:overlap-transform-recursion} with the induction hypothesis gives
\[
\begin{aligned}
    A_Z(2L)
    &\le
    \biggl(\frac nN\biggr)^2 n^{-1/2}
    +
    \biggl(\frac mN\biggr)^2 m^{-1/2}
    +
    \frac{2nm}{N^2}e^{-2L}\\
    &=
    \frac{n^{3/2}+m^{3/2}}{N^2}
    +
    \frac{2nm}{N^4}.
\end{aligned}
\]
Writing $x=n/N$ and $y=m/N$, we have $x+y=1$, and using $N\ge2$,
\[
    \sqrt N\,A_Z(2L)
    \le
    x^{3/2}+y^{3/2}+xy.
\]
Since $x^{3/2}\le (x+x^2)/2$ and $y^{3/2}\le (y+y^2)/2$,
\[
    x^{3/2}+y^{3/2}+xy
    \le
    \frac{x+y+x^2+y^2}{2}+xy
    =
    \frac{1+(x+y)^2}{2}
    =
    1.
\]
Thus \(A_Z(2\log N_Z)\le N_Z^{-1/2}\), completing the induction.
\end{proof}

\begin{proof}[Proof of Theorem~\ref{thm:formula-lb}]
Let $N=N_Z$ and $L=\log N$. The argument below proves the claim for all $N$ larger than a universal constant. The case $N=1$ is trivial, and the finitely many remaining values of \(N\) are absorbed by decreasing the universal constant. Assume therefore that \(N\) is in the large-index regime. By Lemma~\ref{lem:logbalanced-transform} and Markov's inequality,
\[
\begin{aligned}
    \PP\Bigl(D_Z(U,V)\le \frac18\Bigr)
    &=
    \PP\Bigl(
        \exp(-2LD_Z(U,V))\ge \exp(-L/4)
    \Bigr)\\
    &\le
    e^{L/4}\Expt\exp(-2LD_Z(U,V))\\
    &\le
    N^{1/4}N^{-1/2}
    =
    N^{-1/4}.
\end{aligned}
\]
The event $D_Z(U,V)\le1/8$ is the same as $\|\Phi_Z(U)-\Phi_Z(V)\|_2\le1/2$. Build a simple graph \(G_{\mathrm{close}}\) on \(\cI_Z\) by connecting two distinct indices when they are \(1/2\)-close. The number of ordered close pairs is at most \(N^{-1/4}N^2\); after removing diagonal pairs and identifying the two orientations, \(|E(G_{\mathrm{close}})|\le\frac12 N^{-1/4}N^2\).
Lemma~\ref{lem:close-pair-extraction} gives an independent set \(\cJ_Z\subseteq\cI_Z\) with
\[
    |\cJ_Z|
    \ge
    \frac{N^2}{2|E(G_{\mathrm{close}})|+N}
    \ge
    \frac{N^2}{N^{7/4}+N}
    \ge
    c_1N^{1/4}
\]
in the large-index regime. Since \(\cJ_Z\) is independent in \(G_{\mathrm{close}}\), distinct $\iota,\kappa\in\cJ_Z$ satisfy
\[
    \|\Phi_Z(\iota)-\Phi_Z(\kappa)\|_2>\frac12.
\]
Sudakov minoration (Lemma~\ref{lem:sudakov-width-separated}) gives
\[
\begin{aligned}
    \Expt[Z]
    &=
    \Expt\max_{\iota\in\cI_Z}
    \langle \gb_Z,\Phi_Z(\iota)\rangle\\
    &\ge
    \Expt\max_{\iota\in\cJ_Z}
    \langle \gb_Z,\Phi_Z(\iota)\rangle\\
    &\ge
    c_2\sqrt{\log|\cJ_Z|}.
\end{aligned}
\]
Since $|\cJ_Z|\ge c_1N^{1/4}$, this is at least $c\sqrt{\log N}$ after adjusting constants.
\end{proof}

\section{Proof of the Entropy-to-width Bound}
\label{app:effective-entropy}

This section proves the scale-free statement in Theorem~\ref{thm:effective-entropy}. It applies to every tropical Gaussian formula, including formulae whose mix weights are not log-balanced. The only formula-specific input is the Gaussian-process representation from Appendix~\ref{app:formula-proof}; after that, the proof uses only the spherical entropy profile.

For reference, the exact close-pair recursions are as follows. A leaf satisfies $A_G(\lambda)=1$. If $Z=\max\{X,Y\}$, then with $n=N_X,m=N_Y,N=n+m$,
\[
    A_Z(\lambda)
    =
    \biggl(\frac nN\biggr)^2A_X(\lambda)
    +
    \biggl(\frac mN\biggr)^2A_Y(\lambda)
    +
    \frac{2nm}{N^2}e^{-\lambda}.
\]
For a mix node \(Z=\sqrt aX+\sqrt{1-a}Y\),
\[
    A_Z(\lambda)
    =
    A_X(a\lambda)A_Y((1-a)\lambda),
    \qquad
    B_Z(\lambda)
    =
    B_X(a\lambda)+B_Y((1-a)\lambda).
\]
These identities are immediate from the recursive construction of $\Phi_Z$ and from the independence of the two coordinates at a mix node.

\begin{proof}[Proof of Theorem~\ref{thm:effective-entropy}]
Fix $\lambda>0$ with $B_Z(\lambda)\ge2$ and write $B=B_Z(\lambda)$. Let $U,V$ be independent uniformly sampled indices from $\cI_Z$. By Markov's inequality,
\[
\begin{aligned}
    \PP\Bigl(D_Z(U,V)\le \frac{B}{2\lambda}\Bigr)
    &=
    \PP\Bigl(
        \exp(-\lambda D_Z(U,V))\ge\exp(-B/2)
    \Bigr)\\
    &\le
    e^{B/2}\Expt\exp(-\lambda D_Z(U,V))
    =
    e^{-B/2}.
\end{aligned}
\]
Since diagonal pairs have $D_Z(U,V)=0$, we also have $A_Z(\lambda)\ge 1/N_Z$, so $B_Z(\lambda)\le\log N_Z$ and $N_Z\ge e^B$.

Build a simple graph \(G_{\mathrm{close}}\) on \(\cI_Z\) by connecting two distinct indices whenever \(D_Z(\iota,\kappa)\le B/(2\lambda)\). The number of ordered close pairs is at most \(e^{-B/2}N_Z^2\), so
\[
    |E(G_{\mathrm{close}})|
    \le
    \frac12 e^{-B/2}N_Z^2.
\]
Lemma~\ref{lem:close-pair-extraction} gives an independent set \(\cJ_Z\subseteq\cI_Z\) with
\[
    |\cJ_Z|
    \ge
    \frac{N_Z^2}{2|E(G_{\mathrm{close}})|+N_Z}
    \ge
    \frac{N_Z^2}{e^{-B/2}N_Z^2+N_Z}
    \ge
    c_1e^{B/2}.
\]
Here the last inequality uses \(N_Z\ge e^B\).
For distinct $\iota,\kappa\in\cJ_Z$,
\[
    \|\Phi_Z(\iota)-\Phi_Z(\kappa)\|_2
    >
    \sqrt{\frac{B}{\lambda}}.
\]
Sudakov minoration (Lemma~\ref{lem:sudakov-width-separated}) therefore yields
\[
\begin{aligned}
    \Expt[Z]
    &\ge
    c_2
    \sqrt{\frac{B}{\lambda}}
    \sqrt{\log|\cJ_Z|}.
\end{aligned}
\]
Because $|\cJ_Z|\ge c_1e^{B/2}$ and $B\ge2$, $\log|\cJ_Z|\ge c_3B$, and hence
\[
    \Expt[Z]\ge c\frac{B_Z(\lambda)}{\sqrt{\lambda}}.
\]
Taking the supremum over admissible $\lambda$ proves the result.
\end{proof}

\section{Proofs for the Treeplex Reduction}
\label{app:treeplex-reduction}

\begin{proof}[Proof of Lemma~\ref{lem:subtree-plan-counts}]
We argue directly from the definition of reduced normal-form continuation plans. If \(h=\sigma\in\leaveseqs\), there is only the empty continuation below \(\sigma\), hence \(N_\sigma=1\).

If \(h=j\) is a decision point, a pure continuation plan chooses exactly one action \(a\in\cA_j\) and then chooses a pure continuation below \(ja\). These choices form a disjoint union over actions, so
\[
    N_j=\sum_{a\in\cA_j}N_{ja}.
\]

At a nonterminal observation point \(h=\sigma\), a pure continuation plan is a tuple of child continuations, one for each successor \(j\in\cC_\sigma\). Thus \(\cV_\sigma\) is the Cartesian product of the child plan sets, and \(N_\sigma=\prod_{j\in\cC_\sigma}N_j\).
These cases cover all points in \(\tfsdp\).
\end{proof}

\begin{definition}[Log-balanced recursive hard instance]
\label{def:treeplex-hard-instance}
For every subtree \(h\), define a formula \(Z_h\) and a one-round reward process \(W_h:\cV_h\to[-1,1]\) bottom-up as follows.
\begin{itemize}[leftmargin=*]
    \item If \(h=\sigma\) is terminal, \(Z_h\) is a single standard Gaussian and \(W_\sigma(\pi)=\epsilon\), where \(\epsilon\) is Rademacher.
    \item If \(h=j\) is a decision point, set
\[
    Z_j=\max_{a\in\cA_j} Z_{ja},
\]
    where a multiway max may be implemented by any binary bracketing. Independently sample the child one-round processes for all actions, and set \(W_j(\pi)=W_{ja}(\pi_{ja})\) when \(\pi\) chooses action \(a\).
    \item If \(h=\sigma\) is a nonterminal observation point and \(\cC_\sigma=\{j_1,\ldots,j_m\}\), combine the child formulae by repeated log-balanced mix operations. The bracketing-independent form, proved in Lemma~\ref{lem:multiway-log-balanced}, is
\[
    Z_\sigma
    =
    \sum_{r=1}^m \sqrt{p_r}\,Z_{j_r},
    \qquad
    p_r=\frac{\log N_{j_r}}{\sum_{s=1}^m\log N_{j_s}},
\]
    with zero-log subtrees ignored. If all logs vanish, the probabilities are irrelevant. At the same observation point, sample a route \(R\in\{1,\ldots,m\}\) with \(\PP(R=r)=p_r\), route all transition mass to \(j_R\), sample the child one-round process in that subtree, and set \(W_\sigma(\pi)=W_{j_R}(\pi_{j_R})\). Terminal rewards outside the routed subtree are arbitrary in \([-1,1]\), since their transition reach is zero.
\end{itemize}
Across rounds, the hard instance samples independent copies of this recursive construction.
\end{definition}

\begin{lemma}[Index-count preservation]
\label{lem:treeplex-formula-index-count}
The formulae in Definition~\ref{def:treeplex-hard-instance} satisfy
\[
    N_{Z_h}=N_h
    \quad\text{for every }h\in\tfsdp.
\]
\end{lemma}

\begin{proof}[Proof of Lemma~\ref{lem:treeplex-formula-index-count}]
We prove \(N_{Z_h}=N_h\) by induction from the leaves upward. If \(h=\sigma\in\leaveseqs\), then \(Z_\sigma\) is a single Gaussian leaf, so \(N_{Z_\sigma}=1=N_\sigma\).

If \(h=j\) is a decision point, the formula \(Z_j\) is the max over the action-child formulae, implemented by any binary bracketing. Max nodes add index counts, and the induction hypothesis gives
\[
    N_{Z_j}
    =
    \sum_{a\in\cA_j}N_{Z_{ja}}
    =
    \sum_{a\in\cA_j}N_{ja}
    =
    N_j.
\]

If \(h=\sigma\) is a nonterminal observation point, the formula is obtained by repeated mix operations over the child formulae with \(N_{j_r}>1\); child formulae with \(N_{j_r}=1\) may be ignored because they multiply the index count by one. Mix nodes multiply index counts, so the induction hypothesis and Lemma~\ref{lem:subtree-plan-counts} imply \(N_{Z_\sigma}=\prod_{j\in\cC_\sigma}N_{Z_j}=\prod_{j\in\cC_\sigma}N_j=N_\sigma\).
The all-zero-log case is included in the same identity, since then every factor is one.
\end{proof}

\begin{definition}[Centered one-round process]
\label{def:centered-treeplex-process}
For a subtree \(h\) and \(\pi\in\cV_h\), define
\[
\begin{aligned}
    \xi_h(\pi)
    &\triangleq W_h(\pi),\\
    K_h(\pi,\pi')
    &\triangleq
    \Expt[\xi_h(\pi)\xi_h(\pi')].
\end{aligned}
\]
\end{definition}

We next prove the subtree invariant used in Section~\ref{sec:reduction}. Fix a subtree \(h\). For one round, write
\[
    \xi_h(\pi)=W_h(\pi),
    \qquad \pi\in\cV_h.
\]
Then \(\xi_h(\pi)\in[-1,1]\). The distribution of the full vector \(\xi_h=(\xi_h(\pi))_{\pi\in\cV_h}\) is determined by the recursive hard instance.

\begin{lemma}[Multiway log-balanced mix]
\label{lem:multiway-log-balanced}
Assume at least one \(N_{j_r}\) is larger than one. Let \(L_r=\log N_{j_r}\), ignore indices with \(L_r=0\), and let \(L=\sum_r L_r\). Any binary bracketing that repeatedly combines two partial formulae using the log-balanced rule produces
\[
    Z_\sigma=\sum_{r=1}^m\sqrt{L_r/L}\,Z_{j_r}.
\]
Thus the weights \(p_r=L_r/L\) are independent of the bracketing.
\end{lemma}

\begin{proof}
The claim follows by induction over the binary bracketing tree. Suppose a partial block \(S\) has total log-index \(L_S=\sum_{r\in S}L_r\) and contributes coefficient \(\sqrt{L_r/L_S}\) to \(Z_{j_r}\) for \(r\in S\). When blocks \(S\) and \(S'\) are combined, every coefficient from \(S\) is multiplied by \(\sqrt{L_S/(L_S+L_{S'})}\). Hence the coefficient of \(Z_{j_r}\) becomes
\[
    \sqrt{\frac{L_r}{L_S}}\sqrt{\frac{L_S}{L_S+L_{S'}}}
    =
    \sqrt{\frac{L_r}{L_S+L_{S'}}}.
\]
The same calculation applies to \(S'\).
\end{proof}

\subsection{Subtree Invariant}
\label{app:treeplex-realization}

\begin{lemma}[Subtree reduction invariant]
\label{lem:subtree-reduction}
\label{lem:global-realization}
\label{lem:hard-instance-moments}
For every subtree \(h\), the construction above satisfies \(N_{Z_h}=N_h\). For every realization of the recursive one-round process, there exist a valid local transition kernel \(\yb_h\) and terminal reward vector \(\rb_h\in[-1,1]^{\leaveseqs_h}\) with the payoff identity
\[
    W_h(\pi)
    =
    \sum_{\sigma\in\leaveseqs_h}
    \vb_\pi[\sigma]\,\yb_h[\sigma]\,\rb_h[\sigma]
\]
for every pure continuation plan \(\pi\in\cV_h\).
At the root, this gives a feasible full-information reward vector \(\wb=\wb(\yb_\emptyset,\rb_\emptyset)\in\RR^{\leaveseqs}\) with \(\langle \vb_\pi,\wb\rangle=W_\emptyset(\pi)\) for every \(\pi\in\cV\).
Moreover, for every \(\pi\in\cV_h\), \(\Expt[W_h(\pi)]=0\). The kernel \(K_h\) is the covariance kernel generated by the Gaussian-process representation of \(Z_h\): terminal nodes have \(K_\sigma=1\), decision nodes use orthogonal blocks across different actions, and observation nodes use
\[
    K_\sigma(\pi,\pi')
    =
    \sum_{r=1}^m
    p_rK_{j_r}(\pi_{j_r},\pi'_{j_r}).
\]
Consequently \(K_h(\pi,\pi)=1\) for every \(\pi\in\cV_h\). If \(\cG_h=(G_h(\pi))_{\pi\in\cV_h}\) is the centered Gaussian process with covariance \(K_h\), then
\[
    Z_h\stackrel{d}{=}\max_{\pi\in\cV_h}G_h(\pi).
\]
\end{lemma}

\begin{proof}[Proof of Lemma~\ref{lem:subtree-reduction}: feasibility]
We prove the local statement by induction. The local transition kernel is normalized so that the root of the subtree has transition reach one.

If \(h=\sigma\in\leaveseqs\), set \(\yb_\sigma[\sigma]=1\) and \(\rb_\sigma[\sigma]=W_\sigma(\pi)\). This realizes the unique pure continuation reward.

Suppose \(h=j\) is a decision point. For each \(a\in\cA_j\), the recursive construction samples a child realization, and the induction hypothesis gives \((\yb_{ja},\rb_{ja})\) on terminals below \(ja\). Define the local kernel by setting \(\yb_j[j]=1\), \(\yb_j[ja]=1\) for every action \(a\), and by copying \(\yb_{ja}\) inside each child subtree. Define \(\rb_j\) by concatenating the child reward vectors. If \(\pi\in\cV_j\) chooses action \(a\), its terminal indicator has support only below \(ja\), so the inner product against \(\yb_j\odot\rb_j\) equals the child reward \(W_{ja}(\pi_{ja})=W_j(\pi)\).

Finally suppose \(h=\sigma\) is a nonterminal observation point. The construction samples a routed successor \(j_R\). Set \(\yb_\sigma[\sigma]=1\), \(\yb_\sigma[j_R]=1\), and \(\yb_\sigma[j]=0\) for \(j\neq j_R\). Inside the routed subtree, copy the child kernel and reward vector given by induction. Outside it, set all transition coordinates below the unrouted successors to zero and choose arbitrary terminal rewards in \([-1,1]\). Terminals outside \(j_R\) have transition reach zero, while terminals inside \(j_R\) realize \(W_{j_R}(\pi_{j_R})\). Hence every pure continuation plan \(\pi\in\cV_\sigma\) receives \(W_\sigma(\pi)\). Taking \(h=\emptyset\) gives the global reward vector \(\wb(\yb_\emptyset,\rb_\emptyset)\).
\end{proof}

\begin{proof}[Proof of Lemma~\ref{lem:subtree-reduction}: means and covariance]
The identity \(N_{Z_h}=N_h\) is Lemma~\ref{lem:treeplex-formula-index-count}. It remains to prove the zero-mean and covariance claims, which we do by induction from the leaves to the root. If \(h=\sigma\) is terminal, then \(W_\sigma(\pi)=\epsilon\), so \(\Expt[W_\sigma(\pi)]=0\) and \(\Expt[\xi_\sigma(\pi)\xi_\sigma(\pi')]=1\).

Now let \(h=j\) be a decision point. A pure plan \(\pi\in\cV_j\) chooses one action \(a\) and a continuation plan \(\pi_{ja}\in\cV_{ja}\). The construction samples independent child instances for all actions, and
\[
    W_j(\pi)=W_{ja}(\pi_{ja}).
\]
The induction hypothesis gives \(\Expt[W_j(\pi)]=0\). If \(\pi,\pi'\) choose the same action \(a\), their covariance is the child covariance \(K_{ja}(\pi_{ja},\pi'_{ja})\). If they choose different actions, the child variables are independent and have mean zero, so the covariance is \(0\). This is exactly the orthogonal-block recursion of the formula representation.

Finally let \(h=\sigma\) be a nonterminal observation point with successors \(\cC_\sigma=\{j_1,\ldots,j_m\}\). The hard instance first samples a route \(R\) with \(\PP(R=r)=p_r\), then samples the child instance in \(j_R\). For a pure continuation plan \(\pi=(\pi_{j_1},\ldots,\pi_{j_m})\),
\[
    W_\sigma(\pi)=W_{j_R}(\pi_{j_R}),
    \qquad
    \xi_\sigma(\pi)=\xi_{j_R}(\pi_{j_R}).
\]
Conditioning on \(R\) and using the induction hypothesis gives \(\Expt[W_\sigma(\pi)]=0\). For the covariance,
\[
    \Expt[\xi_\sigma(\pi)\xi_\sigma(\pi')]
    =
    \sum_{r=1}^m
    p_rK_{j_r}(\pi_{j_r},\pi'_{j_r}).
\]
This is the same weighted inner-product recursion as the mix node \(\sum_r\sqrt{p_r}Z_{j_r}\). Thus \(K_h(\pi,\pi')\) agrees with \(\langle\Phi_h(\pi),\Phi_h(\pi')\rangle\) for every subtree \(h\). In particular \(K_h(\pi,\pi)=1\). The centered Gaussian process with covariance \(K_h\) has the same finite-dimensional law as the Gaussian process in Lemma~\ref{lem:formula-gp-representation} indexed by \(\cV_h\), and therefore \(Z_h\stackrel{d}{=}\max_{\pi\in\cV_h}G_h(\pi)\).
\end{proof}

\begin{proof}[Proof of Proposition~\ref{prop:treeplex-formula-reduction}]
Apply Definition~\ref{def:treeplex-hard-instance} at the root \(h=\emptyset\). Index-count preservation is Lemma~\ref{lem:treeplex-formula-index-count}; feasibility, centering, covariance matching, and the Gaussian-maximum identity are the root case of Lemma~\ref{lem:subtree-reduction}.
\end{proof}

\section{Rademacher-to-Gaussian Comparison}
\label{app:rademacher-gaussian-comparison}

\begin{corollary}[CCK Gaussian approximation for bounded coordinates]
\label{cor:cck-bounded-coordinates}
Let \(x_1,\ldots,x_n\in\RR^p\) be independent centered random vectors, and set
\[
    X=\frac1{\sqrt n}\sum_{i=1}^n x_i,
    \qquad
    Y\sim\cN\biggl(0,\frac1n\sum_{i=1}^n\Expt[x_ix_i^\top]\biggr).
\]
Let \(T_0=\max_{j\in\llbracket p\rrbracket}X[j]\), \(Z_0=\max_{j\in\llbracket p\rrbracket}Y[j]\), and
\[
    \rho
    =
    \sup_{s\in\RR}
    \bigl|\PP(T_0\le s)-\PP(Z_0\le s)\bigr|.
\]
Assume constants \(0<c_1<C_1\), \(c_2,C_2>0\), and \(B_n\ge1\) satisfy, for every \(j\in\llbracket p\rrbracket\),
\[
    c_1
    \le
    \frac1n\sum_{i=1}^n\Expt[x_i[j]^2]
    \le
    C_1,
\]
and assume the bounded-envelope condition
\[
    \max_{j\in\llbracket p\rrbracket}
    \max_{r=1,2}
    \frac1n\sum_{i=1}^n
    \Expt\bigl[|x_i[j]|^{2+r}/B_n^r\bigr]
    +
    \max_{i\in\llbracket n\rrbracket}
    \max_{j\in\llbracket p\rrbracket}
    \Expt\exp\bigl(|x_i[j]|/B_n\bigr)
    \le4
\]
and the size condition
\[
    \frac{B_n^2\log^7(pn)}{n}
    \le
    C_2n^{-c_2},
\]
hold. Then constants \(C,c>0\), depending only on \(c_1,C_1,c_2,C_2\), satisfy
\[
    \rho\le Cn^{-c}.
\]
\end{corollary}

\begin{proof}
This is the bounded-coordinate specialization of the Gaussian approximation theorem of \citet[Corollary 2.1, derived from Theorem 2.2 and anti-concentration Lemma 2.1]{chernozhukov2013gaussian}, written with the notation used below.
\end{proof}

For \(T\) rounds, let \(W_{t,h}\) be independent copies of \(W_h\), and define
\[
    M_{T,h}
    \triangleq
    \max_{\pi\in\cV_h}
    \sum_{t=1}^T W_{t,h}(\pi).
\]

\begin{lemma}[Finite-time Rademacher-to-Gaussian comparison]
\label{lem:rademacher-gaussian-comparison}
For some universal constants \(C_0,c_0,k>0\), the following holds for every subtree \(h\) with \(N_h\ge2\) and every \(T\ge1\):
\[
    \biggl|
    \Expt\biggl[\frac{M_{T,h}}{\sqrt T}\biggr]
    -
    \Expt[Z_h]
    \biggr|
    \le
    C_0\sqrt{\log(N_h+1)}
    \biggl(\frac{\log^k(N_h+1)}{T}\biggr)^{c_0}.
\]
\end{lemma}

\begin{lemma}[CCK Kolmogorov bound, bounded specialization]
\label{lem:cck-kolmogorov}
Let \(X_1,\ldots,X_T\in\RR^N\) be independent centered random vectors. Assume
\[
    \max_{t\in\llbracket T\rrbracket}
    \max_{i\in\llbracket N\rrbracket}
    |X_t[i]|\le1,
    \qquad
    \frac1T\sum_{t=1}^T\Expt[X_t[i]^2]=1
    \quad \forall i\in\llbracket N\rrbracket.
\]
Let \(\mSigma=T^{-1}\sum_{t=1}^T\Cov(X_t)\) and \(G\sim\cN(0,\mSigma)\). For some universal constants \(C,c,k>0\),
\[
    \sup_{s\in\RR}
    \biggl|
    \PP\biggl(
    \max_{i\in\llbracket N\rrbracket}
    \frac1{\sqrt T}\sum_{t=1}^T X_t[i]\le s
    \biggr)
    -
    \PP\biggl(
    \max_{i\in\llbracket N\rrbracket}G[i]\le s
    \biggr)
    \biggr|
    \le
    C
    \biggl(\frac{\log^k(N+1)}{T}\biggr)^c.
\]
\end{lemma}

\begin{proof}
We spell out the substitution into Corollary~\ref{cor:cck-bounded-coordinates}. The CCK sample size and dimension are
\[
    n=T,
    \qquad
    p=N+1.
\]
The extra coordinate is a duplicate of the first coordinate, which leaves both maxima unchanged. In CCK notation \(x_{t,i}=X_t[i]\) for \(i\le N\) and \(x_{t,N+1}=X_t[1]\). Their statistic \(T_0\) is the maximum of \(T^{-1/2}\sum_{t=1}^T X_t\), and their \(Z_0\) is the covariance-matched Gaussian maximum.

Condition (E.1) of CCK holds with \(B_n=1\). Indeed,
\[
    \frac1T\sum_{t=1}^T\Expt[x_{t,i}^2]
    =
    1,
\]
so one may take their variance constants to be \(c_1=1/2\) and \(C_1=2\). Moreover \(|x_{t,i}|\le1\), so the moment term is at most \(1\) for \(r=1,2\), and \(\Expt[\exp(|x_{t,i}|/B_n)]\le e\). Together these bounds verify the CCK bounded-envelope condition:
\[
    \max_{r=1,2}\frac1T\sum_{t=1}^T\Expt[|x_{t,i}|^{2+r}/B_n^r]
    +\Expt[\exp(|x_{t,i}|/B_n)]\le4.
\]

Fix the CCK polynomial-rate parameters \(c_2=1/2\) and \(C_2=1\). If
\[
    \frac{\log^7((N+1)T)}{T}\le T^{-1/2},
\]
then CCK Corollary 2.1 gives
\[
    \Delta_{\mathrm{K}}
    \triangleq
    \sup_{s\in\RR}
    \bigl|\PP(T_0\le s)-\PP(Z_0\le s)\bigr|
    \le
    C_{\mathrm{CCK}}T^{-c_{\mathrm{CCK}}},
\]
where \(C_{\mathrm{CCK}},c_{\mathrm{CCK}}>0\) depend only on the fixed constants \(1/2,2,1/2,1\), hence are universal.
Since \(L\triangleq\log(N+1)\ge\log2\), this implies
\[
    \Delta_{\mathrm{K}}
    \le
    C_{\mathrm{CCK}}(\log2)^{-14c_{\mathrm{CCK}}}
    \biggl(\frac{L^{14}}{T}\biggr)^{c_{\mathrm{CCK}}}.
\]

It remains only to remove \(\log T\) from the displayed condition. If the condition fails, then \(T<\log^{14}((N+1)T)\). Since
\[
    \log((N+1)T)=L+\log T,
\]
either \(\log T\le L\), in which case \(T<(2L)^{14}\), or \(\log T>L\), in which case \(T<(2\log T)^{14}\). Set
\[
    C_\dagger
    \triangleq
    \sup\{u\ge1:u\le(2\log u)^{14}\}<\infty.
\]
In the second case \(T\le C_\dagger\le C_\dagger(\log2)^{-14}L^{14}\). Hence, in both cases \(T\le C_*L^{14}\) for a universal \(C_*\). Thus, when the CCK condition fails,
\[
    1\le C_*^c
    \biggl(\frac{\log^{14}(N+1)}{T}\biggr)^c
\]
for every fixed \(c>0\). Combining this trivial estimate with the CCK estimate in the complementary case gives the claim with \(k=14\) and \(c=c_{\mathrm{CCK}}\) after increasing \(C\).
\end{proof}

\begin{lemma}[Bounded high-dimensional CLT in expectation]
\label{lem:cck-expectation}
Let \(X_1,\ldots,X_T\in\RR^N\) be independent centered random vectors. Assume
\[
    \max_{t\in\llbracket T\rrbracket}
    \max_{i\in\llbracket N\rrbracket}
    |X_t[i]|\le1,
    \qquad
    \frac1T\sum_{t=1}^T\Expt[X_t[i]^2]=1
    \quad \forall i\in\llbracket N\rrbracket.
\]
Let \(\mSigma=T^{-1}\sum_{t=1}^T\Cov(X_t)\) and \(G\sim\cN(0,\mSigma)\). For some universal constants \(C,c,k>0\),
\[
    \biggl|
    \Expt\max_{i\in\llbracket N\rrbracket}
    \frac1{\sqrt T}\sum_{t=1}^T X_t[i]
    -
    \Expt\max_{i\in\llbracket N\rrbracket}G[i]
    \biggr|
    \le
    C\sqrt{\log(N+1)}
    \biggl(\frac{\log^k(N+1)}{T}\biggr)^c.
\]
\end{lemma}

\begin{proof}
Define the two maxima
\[
    S=\max_{i\in\llbracket N\rrbracket}
    \frac1{\sqrt T}\sum_{t=1}^T X_t[i],
    \qquad
    Z=\max_{i\in\llbracket N\rrbracket}G[i].
\]
By Lemma~\ref{lem:cck-kolmogorov}, the Kolmogorov distance
\[
    \Delta
    \triangleq
    \sup_{s\in\RR}
    \bigl|\PP(S\le s)-\PP(Z\le s)\bigr|
    \le
    C_1
    \biggl(\frac{\log^k(N+1)}{T}\biggr)^{c_1}
\]
for universal \(C_1,c_1>0\). If the right-hand side is larger than \(1/2\), the final expectation bound follows from the elementary estimate \(\Expt|S|+\Expt|Z|\le C\sqrt{\log(N+1)}\) after increasing \(C\). If \(\Delta=0\), there is nothing to prove, so assume below that \(0<\Delta\le1/2\).

The smoothing and Gaussian anti-concentration steps are contained entirely in the CCK Kolmogorov bound. The remaining step here is the expectation conversion.

For \(R>0\), the tail-integral identity gives
\[
    |\Expt S-\Expt Z|
    \le
    2R\Delta
    +
    \Expt\bigl[(|S|-R)_+\bigr]
    +
    \Expt\bigl[(|Z|-R)_+\bigr].
\]
Hoeffding's inequality and a union bound give
\[
    \PP(|S|>u)\le 2(N+1)e^{-u^2/2},
\]
and the same bound holds for \(Z\) by the Gaussian tail bound and \(\mSigma_{ii}=1\). Choose
\[
    R=2\sqrt{\log(N+1)}+\sqrt{2\log(1/\Delta)}.
\]
The two tail integrals are then at most \(C_2\Delta\). Therefore
\[
    |\Expt S-\Expt Z|
    \le
    C_3\Delta
    \Bigl(\sqrt{\log(N+1)}+\sqrt{\log(1/\Delta)}+1\Bigr).
\]
Let \(A=\log^k(N+1)/T\). Since \(\sqrt{\log(N+1)}\ge\sqrt{\log2}\) and \(\Delta\sqrt{\log(1/\Delta)}\le C_4\Delta^{1/2}\) for \(\Delta\in(0,1/2]\), the last display is at most
\[
    C_5\sqrt{\log(N+1)}\,\Delta^{1/2}
    \le
    C_6\sqrt{\log(N+1)}\,A^{c_1/2}.
\]
Thus the expectation comparison holds with exponent \(c=c_1/2\).
\end{proof}

\begin{proof}[Proof of Lemma~\ref{lem:rademacher-gaussian-comparison}]
For fixed \(h\), the set \(\cV_h\) is finite. Let \(N=N_h\), and write
\[
    X_t
    \triangleq
    \bigl(\xi_{t,h}(\pi)\bigr)_{\pi\in\cV_h},
    \qquad
    \xi_{t,h}(\pi)=W_{t,h}(\pi),
\]
where \(X_1,\ldots,X_T\) are independent copies of the one-round vector. By Lemma~\ref{lem:subtree-reduction}, these vectors have mean zero, covariance matrix \(K_h\), and unit coordinate variances. Also \(|X_t[\pi]|\le1\). The random route at an observation point is already part of the one-round law, so no separate conditioning on route counts is needed.

Applying Lemma~\ref{lem:cck-expectation} with coordinates indexed by \(\pi\in\cV_h\) gives
\[
    \Expt\max_{\pi\in\cV_h}
    \frac{1}{\sqrt T}\sum_{t=1}^T \xi_{t,h}(\pi)
    =
    \Expt\biggl[\frac{M_{T,h}}{\sqrt T}\biggr]
\]
up to the displayed error. The covariance-matched Gaussian maximum is
\[
    \Expt\max_{\pi\in\cV_h}G_h(\pi)=\Expt[Z_h],
\]
by Lemma~\ref{lem:subtree-reduction}. The result is the displayed comparison bound after renaming constants.
\end{proof}

\section{Proof of the EFG Lower Bound}
\label{app:efg-lower-bound}

\begin{proof}[Proof of Theorem~\ref{thm:efg-lb}]
Apply the recursive construction at the root \(h=\emptyset\), and let \(N=|\cV|\). If \(N=1\), the lower bound is vacuous. Assume \(N\ge2\). By Lemma~\ref{lem:subtree-reduction}, each one-round draw is realized by a valid transition kernel and terminal reward vector. The same lemma gives \(N_{Z_\emptyset}=|\cV|\) and zero one-round mean for every pure plan.

Fix an arbitrary online learner and condition on the history before round \(t\). By Lemma~\ref{lem:treeplex-conv}, the learner's chosen strategy \(\xb_t\in\cQ\) is a convex combination of pure plans. Since every pure plan has conditional expected reward \(0\), the learner's conditional expected reward is also \(0\). Hence the learner's expected cumulative reward is \(0\).

The best fixed linear comparator over \(\cQ=\co(\cV)\) is attained at a pure plan. Its reward is
\[
    \max_{\pi\in\cV}
    \sum_{t=1}^T W_{t,\emptyset}(\pi)
    =
    M_{T,\emptyset},
\]
so the expected regret is at least \(\Expt[M_{T,\emptyset}]\). Lemma~\ref{lem:rademacher-gaussian-comparison} gives
\[
    \Expt[M_{T,\emptyset}]
    \ge
    \sqrt T
    \biggl(
    \Expt[Z_\emptyset]
    -
    C_0\sqrt{\log(N+1)}
    \biggl(\frac{\log^k(N+1)}{T}\biggr)^{c_0}
    \biggr).
\]
The formula lower bound gives
\[
    \Expt[Z_\emptyset]\ge c_1\sqrt{\log N}.
\]
For \(N\ge2\), \(\sqrt{\log(N+1)}\le c_2\sqrt{\log N}\). Here \(k=14\), as in the proof of Lemma~\ref{lem:cck-kolmogorov}. Choosing the universal horizon constant \(C\) sufficiently large makes the comparison error at most \(c_1\sqrt{\log N}/2\). Therefore
\[
    \Expt[\Regret(T)]
    \ge
    c\sqrt{T\log|\cV|}
\]
for a possibly smaller universal constant \(c>0\).
\end{proof}

\paragraph{Remarks.}
We remark that the treeplex is fixed first: the hard environment, a fixed routing distribution held across all \(T\) rounds together with independent Rademacher terminal rewards, is then chosen compatibly with it. This environment is a valid full-information EFG reward process that realizes the tropical Gaussian formula.

On the simplex, independent Rademacher rewards already give the \(\sqrt{T\log K}\) lower bound: the learner has zero expected reward in every round, while the best vertex accumulates a positive reward in hindsight. Theorem~\ref{thm:efg-lb} shows that this mechanism survives the much stronger correlations induced by a treeplex. The environment is oblivious and draws independent rewards across rounds, so no adaptivity, temporal dependence, or pathological payoff is needed.

\section{Proof of DilEnt Certificate Optimality}
\label{app:dilent-optimality}

\begin{proof}[Proof of Corollary~\ref{cor:dil-ent-optimal}]
Suppose, toward contradiction, that a regularizer-and-norm pair with uniform feasible-reward control had \(D_\varphi/\mu=o(\log|\cV|)\) along a sequence of treeplexes. Optimizing the standard OMD guarantee over \(\eta\) would give regret
\[
    \Regret(T)
    \le
    \cO\bigl(\sqrt{D_\varphi T/\mu}\bigr)
    =
    o\bigl(\sqrt{T\log|\cV|}\bigr)
\]
uniformly over the feasible reward sequences used in Theorem~\ref{thm:efg-lb}. Taking \(T\) above the polylogarithmic threshold in that theorem contradicts the hard-instance lower bound. Hence every such certificate must satisfy \(D_\varphi/\mu\ge\Omega(\log|\cV|)\). The matching DilEnt upper bound is exactly the treeplex-norm fact stated in Section~\ref{sec:prelim}: weight-one DilEnt is \(1\)-strongly convex under \(\|\cdot\|\Ynorm\), its relevant Bregman diameter is at most \(\log|\cV|\), and \(\|\cdot\|\Qnorm\) controls all feasible reward vectors.
\end{proof}

\section{Fixed Observation Distributions}
\label{app:arbitrary-routing}

\begin{corollary}[Fixed observation outcome distribution lower bound]
\label{cor:arbitrary-routing}
Fix, at each observation point, a distribution over routed successors, and let \(Z_\emptyset\) be the tropical Gaussian formula obtained by using the corresponding mix weights. Along an oblivious feasible reward process induced by bounded terminal rewards and transition outcome sampled from those distributions,
\[
    \Expt[\Regret(T)]
    \ge
    \sqrt T
    \biggl(
    c\sqrt{\Heff(Z_\emptyset)}
    -
    C_0\sqrt{\log(|\cV|+1)}
    \biggl(\frac{\log^k(|\cV|+1)}{T}\biggr)^{c_0}
    \biggr).
\]
This bound holds for every online learning algorithm.
\end{corollary}

\begin{proof}
Run the same hard instance with the prescribed observation outcome distributions. The subtree invariant and the finite-time Gaussian comparison remain valid with those probabilities. As in the proof of Theorem~\ref{thm:efg-lb}, expected regret is at least the expected Rademacher comparator maximum. By Lemma~\ref{lem:rademacher-gaussian-comparison} and Theorem~\ref{thm:effective-entropy},
\[
    \Expt[Z_\emptyset]\ge c\sqrt{\Heff(Z_\emptyset)}.
\]
The displayed bound follows from the same comparison argument.
\end{proof}

If some observation outcome probabilities are fixed in advance, the corresponding mix nodes keep those weights, and the same bound applies. This corollary keeps two quantities separate. The treeplex fixes the index count, so \(N_{Z_\emptyset}=|\cV|\). The observation outcome distribution determines the spherical entropy profile \(B_Z\). If a large subtree receives very small routing probability, its pure plans still contribute to \(N_Z\), but they contribute little to \(B_Z(\lambda)\) at the relevant resolution. The log-balanced hard instance in Theorem~\ref{thm:efg-lb} chooses the observation outcome distribution so that this profile has order \(\log|\cV|\) at the root.

Corollary~\ref{cor:arbitrary-routing} concerns a distribution over observation outcomes, not a single fractional transition kernel fixed across the round. With a fixed fractional kernel, a pure plan reward after an observation node is typically a weighted sum over all successor branches, and the covariance recursion generally changes, for example to a squared-weight form, unless the model also observes realized routes or rescales terminal rewards. That fixed-fractional setting is therefore outside the corollary.

\section{Worked Example}
\label{app:worked-example}

We instantiate the main objects on the treeplex of \Cref{fig:efg-formula-bridge}. The decision point \(j\) has actions \(L\) and \(R\). The sequence \(\sigma=jR\) is an observation point with successors \(\cC_\sigma=\{j_1,j_2\}\), where \(j_1\) has actions \(A,B\) and \(j_2\) has actions \(C,D\). The terminal sequences, in this order, are \(jL,j_1A,j_1B,j_2C,j_2D\).

\paragraph{Pure plans.}
Lemma~\ref{lem:subtree-plan-counts} gives \(N_{j_1}=N_{j_2}=2\), \(N_\sigma=4\), and \(N_j=5\). The pure plans are \(L\) and \((R,x,y)\) for \(x\in\{A,B\}\) and \(y\in\{C,D\}\). Take \(\pi=(R,A,D)\). Its extended sequence-form vector equals one on \(\emptyset,j,jR,j_1,j_1A,j_2,j_2D\) and zero elsewhere, so its terminal projection is \(\vb_\pi=(0,1,0,0,1)\).

\paragraph{Transition kernel and reward.}
Since \(\log N_{j_1}=\log N_{j_2}\), log-balanced routing samples \(r\in\{1,2\}\) uniformly at \(\sigma\). The kernel \(\yb\) routes all mass to \(j_r\): it equals one on \(\emptyset,j,jL,jR,j_r\) and on both actions at \(j_r\), and zero below the other successor. With independent Rademacher variables \(\epsilon_L,\epsilon_A,\epsilon_B,\epsilon_C,\epsilon_D\) as terminal rewards, the reward vector is \(\wb=(\epsilon_L,\epsilon_A,\epsilon_B,0,0)\) when \(r=1\) and \(\wb=(\epsilon_L,0,0,\epsilon_C,\epsilon_D)\) when \(r=2\). Hence \(W(\pi)=\langle\vb_\pi,\wb\rangle\) equals \(\epsilon_A\) or \(\epsilon_D\) with equal probability.

\paragraph{Covariance and formula.}
Every plan has mean zero and variance one. Two plans sharing exactly one of \(x,y\) agree on exactly one route, so their covariance is \(1/2\). The remaining pairs of distinct plans are uncorrelated. This is the covariance of
\[
Z_j
=
\max\Bigl\{G_L,\ \tfrac{1}{\sqrt2}\max\{G_A,G_B\}+\tfrac{1}{\sqrt2}\max\{G_C,G_D\}\Bigr\}.
\]
In Lemma~\ref{lem:formula-gp-representation}, with coordinates \((L,A,B,C,D)\), we have \(\Phi(L)=e_L\) and \(\Phi(R,x,y)=(e_x+e_y)/\sqrt2\).

\paragraph{Spherical entropy.}
Among the \(25\) ordered pairs of indices, \(5\) have \(D_Z=0\), \(8\) have \(D_Z=1/2\), and \(12\) have \(D_Z=1\). Therefore
\[
A_Z(\lambda)=\frac{5+8e^{-\lambda/2}+12e^{-\lambda}}{25},
\]
which agrees with the recursion of Lemma~\ref{lem:overlap-transform-recursion}. At \(\lambda=2\log5\), this gives \(A_Z(\lambda)=0.2832\le5^{-1/2}\), as Lemma~\ref{lem:logbalanced-transform} predicts. For larger formulae, the same scale feeds Theorem~\ref{thm:effective-entropy} and yields Theorem~\ref{thm:formula-lb}.

\newpage
\input{checklist.tex}

\end{document}

%% file: figures/efg_formula_bridge.tex
\begingroup
\tikzset{
    toy panel/.style={
        draw=black!12,
        fill=black!1,
        rounded corners=2pt,
        line width=.45pt
    },
    toy title/.style={
        font=\bfseries\small,
        align=center
    },
    toy label/.style={
        font=\footnotesize,
        inner sep=1pt
    },
    toy small label/.style={
        font=\scriptsize,
        inner sep=.8pt
    },
    toy edge label/.style={
        font=\scriptsize,
        inner sep=.6pt
    },
    toy action/.style={
        semithick,
        black,
        -{Stealth[scale width=1.05,inset=.25mm,length=1.55mm]}
    },
    toy observation/.style={
        semithick,
        black,
        dotted,
        -{Stealth[scale width=1.05,inset=.25mm,length=1.55mm]}
    },
    toy selected action/.style={
        line width=.9pt,
        fancyBlue,
        -{Stealth[scale width=1.1,inset=.25mm,length=1.6mm]}
    },
    toy selected observation/.style={
        line width=.9pt,
        fancyBlue,
        dotted,
        -{Stealth[scale width=1.1,inset=.25mm,length=1.6mm]}
    },
    toy decision/.style={
        draw=black,
        fill=black,
        rounded corners=.35pt,
        minimum size=5pt,
        inner sep=0pt
    },
    toy obs/.style={
        draw=black,
        fill=white,
        circle,
        semithick,
        inner sep=1.15pt,
        path picture={
            \draw[black,line width=.3pt]
                (path picture bounding box.south east) --
                (path picture bounding box.north west)
                (path picture bounding box.south west) --
                (path picture bounding box.north east);
        }
    },
    toy terminal/.style={
        draw=black,
        fill=white,
        minimum size=5pt,
        inner sep=0pt
    },
    toy vertex/.style={
        draw=black!25,
        fill=white,
        rounded corners=1.4pt,
        minimum width=1.18cm,
        minimum height=.33cm,
        inner xsep=2pt,
        inner ysep=1.4pt,
        font=\scriptsize
    },
    toy selected vertex/.style={
        toy vertex,
        draw=fancyBlue,
        fill=fancyBlue!8,
        line width=.75pt
    },
    toy formula/.style={
        draw=black!20,
        fill=white,
        rounded corners=1.5pt,
        inner xsep=4pt,
        inner ysep=5pt,
        font=\scriptsize,
        align=center
    }
}

\begin{tikzpicture}[x=1cm,y=1cm,line cap=round,line join=round]
    \filldraw[toy panel] (-.35,.05) rectangle (4.35,5.15);
    \filldraw[toy panel] (4.75,.05) rectangle (9.45,5.15);
    \filldraw[toy panel] (9.85,.05) rectangle (15.65,5.15);

    \node[toy title] at (2.0,4.85) {tree-form decision problem};
    \node[toy decision] (j) at (2.0,4.05) {};
    \node[toy terminal] (lterm) at (.85,3.2) {};
    \node[toy obs] (sigma) at (3.1,3.2) {};
    \node[toy decision] (jone) at (2.45,2.2) {};
    \node[toy decision] (jtwo) at (3.65,2.2) {};
    \node[toy terminal] (a) at (2.05,1.28) {};
    \node[toy terminal] (b) at (2.85,1.28) {};
    \node[toy terminal] (c) at (3.25,1.28) {};
    \node[toy terminal] (d) at (4.05,1.28) {};

    \draw[toy action] (j) -- (lterm);
    \draw[toy action] (j) -- (sigma);
    \draw[toy observation] (sigma) -- (jone);
    \draw[toy observation] (sigma) -- (jtwo);
    \draw[toy action] (jone) -- (a);
    \draw[toy action] (jone) -- (b);
    \draw[toy action] (jtwo) -- (c);
    \draw[toy action] (jtwo) -- (d);

    \draw[toy selected action] (j) -- (sigma);
    \draw[toy selected observation] (sigma) -- (jone);
    \draw[toy selected observation] (sigma) -- (jtwo);
    \draw[toy selected action] (jone) -- (a);
    \draw[toy selected action] (jtwo) -- (d);

    \node[toy edge label] at (1.35,3.78) {$L$};
    \node[toy edge label] at (2.58,3.78) {$R$};
    \node[toy edge label] at (2.7,2.82) {$p_1$};
    \node[toy edge label] at (3.5,2.82) {$p_2$};
    \node[toy edge label] at (2.02,1.73) {$A$};
    \node[toy edge label] at (2.88,1.73) {$B$};
    \node[toy edge label] at (3.27,1.73) {$C$};
    \node[toy edge label] at (4.05,1.73) {$D$};

    \node[toy label,left=1pt of j] {$j$};
    \node[toy label,right=1pt of sigma] {$\sigma$};
    \node[toy label,left=1pt of jone] {$j_1$};
    \node[toy label,right=1pt of jtwo] {$j_2$};
    \node[toy small label] at (2.0,.35) {$\cV_j=\{L\}\cup\bigl(\cV_{j_1}\times\cV_{j_2}\bigr)$};

    \node at (4.55,2.6) {$\Longrightarrow$};

    \node[toy title] at (7.1,4.85) {pure-plan vertices};
    \node[toy vertex] at (7.1,4.02) {$L$};
    \node[toy vertex] at (6.25,3.4) {$(R,A,C)$};
    \node[toy selected vertex] at (7.95,3.4) {$(R,A,D)$};
    \node[toy vertex] at (6.25,2.78) {$(R,B,C)$};
    \node[toy vertex] at (7.95,2.78) {$(R,B,D)$};

    \node[toy formula,text width=3.9cm] at (7.1,1.58) {$
        \begin{aligned}
            \cV_{j_1}&=\{A,B\},\\
            \cV_{j_2}&=\{C,D\},\\
            N_j&=1+2\cdot2=5.
        \end{aligned}
    $};
    \node[toy small label,align=center] at (7.1,.35) {highlighted vertex: \((R,A,D)\)};

    \node at (9.65,2.6) {$\Longrightarrow$};

    \node[toy title] at (12.75,4.85) {tropical Gaussian formula};
    \node[toy formula,text width=5.0cm] at (12.75,3.35) {$
        \begin{aligned}
            Z_j&=\max\{G_L,Z_\sigma\},\\
            Z_\sigma&=\frac{1}{\sqrt2}Z_{j_1}
                +\frac{1}{\sqrt2}Z_{j_2},\\
            Z_{j_1}&=\max\{G_A,G_B\},\\
            Z_{j_2}&=\max\{G_C,G_D\}.
        \end{aligned}
    $};
    \node[toy formula,text width=5.0cm] at (12.75,1.15) {$
        \begin{aligned}
        Z_j=\max\Bigl\{&G_L,\,
        \frac{G_A+G_C}{\sqrt2},\,
        \textcolor{fancyBlue}{\frac{G_A+G_D}{\sqrt2}},\\
        &\frac{G_B+G_C}{\sqrt2},\,
        \frac{G_B+G_D}{\sqrt2}\Bigr\}.
        \end{aligned}
    $};
\end{tikzpicture}
\endgroup

%% file: templates/checklist.tex
\section*{NeurIPS Paper Checklist}

\begin{enumerate}

\item {\bf Claims}
    \item[] Question: Do the main claims made in the abstract and introduction accurately reflect the paper's contributions and scope?
    \item[] Answer: \answerYes{}%
    \item[] Justification: The claims made in the abstract and introduction accurately reflect the paper's contributions and scope.

\item {\bf Limitations}
    \item[] Question: Does the paper discuss the limitations of the work performed by the authors?
    \item[] Answer: \answerYes{}%
    \item[] Justification: The horizon requirement of the lower bound is stated in the Introduction and in Theorem~\ref{thm:efg-lb}.

\item {\bf Theory assumptions and proofs}
    \item[] Question: For each theoretical result, does the paper provide the full set of assumptions and a complete (and correct) proof?
    \item[] Answer: \answerYes{}%
    \item[] Justification: The theoretical results are all proved.

    \item {\bf Experimental result reproducibility}
    \item[] Question: Does the paper fully disclose all the information needed to reproduce the main experimental results of the paper to the extent that it affects the main claims and/or conclusions of the paper (regardless of whether the code and data are provided or not)?
    \item[] Answer: \answerNA{}%
    \item[] Justification: The paper does not include experiments.

\item {\bf Open access to data and code}
    \item[] Question: Does the paper provide open access to the data and code, with sufficient instructions to faithfully reproduce the main experimental results, as described in supplemental material?
    \item[] Answer: \answerNA{}%
    \item[] Justification: The paper does not include experiments requiring code.

\item {\bf Experimental setting/details}
    \item[] Question: Does the paper specify all the training and test details (e.g., data splits, hyperparameters, how they were chosen, type of optimizer) necessary to understand the results?
    \item[] Answer: \answerNA{}%
    \item[] Justification: The paper does not include experiments.

\item {\bf Experiment statistical significance}
    \item[] Question: Does the paper report error bars suitably and correctly defined or other appropriate information about the statistical significance of the experiments?
    \item[] Answer: \answerNA{}%
    \item[] Justification: The paper does not include experiments.

\item {\bf Experiments compute resources}
    \item[] Question: For each experiment, does the paper provide sufficient information on the computer resources (type of compute workers, memory, time of execution) needed to reproduce the experiments?
    \item[] Answer: \answerNA{}%
    \item[] Justification: The paper does not include experiments.
    
\item {\bf Code of ethics}
    \item[] Question: Does the research conducted in the paper conform, in every respect, with the NeurIPS Code of Ethics \url{https://neurips.cc/public/EthicsGuidelines}?
    \item[] Answer: \answerYes{}%
    \item[] Justification: Yes

\item {\bf Broader impacts}
    \item[] Question: Does the paper discuss both potential positive societal impacts and negative societal impacts of the work performed?
    \item[] Answer: \answerNA{}%
    \item[] Justification: There is no societal impact of the work performed.
    
\item {\bf Safeguards}
    \item[] Question: Does the paper describe safeguards that have been put in place for responsible release of data or models that have a high risk for misuse (e.g., pre-trained language models, image generators, or scraped datasets)?
    \item[] Answer: \answerNA{}%
    \item[] Justification: The paper poses no such risks.

\item {\bf Licenses for existing assets}
    \item[] Question: Are the creators or original owners of assets (e.g., code, data, models), used in the paper, properly credited and are the license and terms of use explicitly mentioned and properly respected?
    \item[] Answer: \answerNA{}%
    \item[] Justification: The paper does not use existing assets.

\item {\bf New assets}
    \item[] Question: Are new assets introduced in the paper well documented and is the documentation provided alongside the assets?
    \item[] Answer: \answerNA{}%
    \item[] Justification: The paper does not release new assets.

\item {\bf Crowdsourcing and research with human subjects}
    \item[] Question: For crowdsourcing experiments and research with human subjects, does the paper include the full text of instructions given to participants and screenshots, if applicable, as well as details about compensation (if any)? 
    \item[] Answer: \answerNA{}%
    \item[] Justification: The paper does not involve crowdsourcing nor research with human subjects.

\item {\bf Institutional review board (IRB) approvals or equivalent for research with human subjects}
    \item[] Question: Does the paper describe potential risks incurred by study participants, whether such risks were disclosed to the subjects, and whether Institutional Review Board (IRB) approvals (or an equivalent approval/review based on the requirements of your country or institution) were obtained?
    \item[] Answer: \answerNA{}%
    \item[] Justification: The paper does not involve crowdsourcing nor research with human subjects.

\item {\bf Declaration of LLM usage}
    \item[] Question: Does the paper describe the usage of LLMs if it is an important, original, or non-standard component of the core methods in this research? Note that if the LLM is used only for writing, editing, or formatting purposes and does \emph{not} impact the core methodology, scientific rigor, or originality of the research, declaration is not required.
    \item[] Answer: \answerYes{}%
    \item[] Justification: The authors proved the main theorem in October 2025. The final proof uses optimizations suggested by an LLM, and several computations in the technical lemma were first derived by an LLM.

\end{enumerate}